\documentclass[preprint,12pt]{elsarticle}

\usepackage{amssymb}
\usepackage{amsmath}
\usepackage{multirow}
\usepackage{makecell}
\usepackage{booktabs}
\usepackage{xcolor}

\journal{Robotics and Autonomous Systems}

\begin{document}

\begin{frontmatter}



\title{Generalizable task-oriented object grasping through LLM-guided ontology and similarity-based planning}


\author[label1]{Hao Chen\corref{cor1}}
\ead{hchenrsjp@gmail.com}
\author[label1]{Takuya Kiyokawa}
\author[label1]{Weiwei Wan}
\author[label1]{Kensuke Harada}

\affiliation[label1]{organization={Graduate School of Engineering Science, The University of Osaka},
            addressline={1-3, Machikaneyamacho, Toyonaka},
            city={Osaka},
            postcode={560-8531},
            country={Japan}}

\cortext[cor1]{Corresponding author}

\begin{abstract}
Task-oriented grasping (TOG) is more challenging than simple object grasping because it requires precise identification of object parts and careful selection of grasping areas to ensure effective and robust manipulation. While recent approaches have trained large-scale vision-language models to integrate part-level object segmentation with task-aware grasp planning, their instability in part recognition and grasp inference limits their ability to generalize across diverse objects and tasks. To address this issue, we introduce a novel, geometry-centric strategy for more generalizable TOG that does not rely on semantic features from visual recognition, effectively overcoming the viewpoint sensitivity of model-based approaches. Our main proposals include: 1) an object-part-task ontology for functional part selection based on intuitive human commands, constructed using a Large Language Model (LLM); 2) a sampling-based geometric analysis method for identifying the selected object part from observed point clouds, incorporating multiple point distribution and distance metrics; and 3) a similarity matching framework for imitative grasp planning, utilizing similar known objects with pre-existing segmentation and grasping knowledge as references to guide the planning for unknown targets. We validate the high accuracy of our approach in functional part selection, identification, and grasp generation through real-world experiments. Additionally, we demonstrate the method's generalization capabilities to novel-category objects by extending existing ontological knowledge, showcasing its adaptability to a broad range of objects and tasks.
\end{abstract}


\begin{highlights}
\item Large language models with ontological knowledge are integrated to guide task-oriented object grasping.
\item We propose a similarity-based geometric analysis method for precise part-level object recognition.
\item Model templates with prior knowledge are utilized for object part matching and imitative grasp planning.
\item We present a novel local-to-global point cloud registration technique for robust grasp knowledge transfer.
\item We introduce a stability-aware positional adjustment strategy for grasp optimization.
\end{highlights}

\begin{keyword}
Task-oriented grasping \sep Ontology \sep Planning under uncertainty


\end{keyword}

\end{frontmatter}



\section{Introduction}
Robotic grasping is advancing beyond conventional pick-and-place operations toward more human-like behaviors, where understanding \textit{part affordance} \cite{Deng,Li} is crucial for achieving task-oriented grasping (TOG). For instance, a robot is expected to identify and grasp the \textit{handle} of a cup in a \textit{pouring} task. In real-world scenarios, however, human instructions are often intuitive and semantically rich, extending far beyond simple action labels like \textit{pouring}. As a result, the accurate interpretation of user intentions becomes essential for both human-guided manipulation and human-robot collaboration tasks. Moreover, the reliable recognition of object functional parts from partial observation while ensuring the generation of high-quality grasps remains a significant challenge, leaving this research problem largely unsolved.

More than a decade ago, the concept of \textit{semantic grasping} was introduced \cite{Dang,Song} to enable high-level manipulation considering task constraints, rather than focusing solely on basic grasping. However, at the time, even with the aid of vision and tactile sensors, analytical approaches struggled to generate high-quality grasps without human supervision and showed limited generalization. With the rapid progress in deep learning and image segmentation, more advanced strategies based on visual recognition have emerged \cite{Detry,Chu,Murali}, demonstrating notable performance in planning task-oriented grasps across a wide variety of objects. Nevertheless, these vision-based methods typically represent target tasks using predefined cues such as \textit{grasp from the handle} or simplistic labels like \textit{handover} or \textit{cut}, lacking the ability to translate contextual human instructions into specific robotic actions. This interaction challenge has been widely addressed in recent years with the advent of vision-language models (VLMs) \cite{Tang,Wei,Zhang}. By leveraging high-dimensional language embeddings and attention-based learning frameworks, robots are now increasingly capable of responding to natural human commands. However, several limitations persist in current research: 1) low grasping DoF \cite{Tang}; 2) reliance on complete mesh models or point clouds \cite{Wei}; and 3) significant performance degradation when generalizing to unseen objects \cite{Zhang}.

To overcome these limitations, we propose a novel strategy for TOG that integrates an object-part-task ontology guided by large language models (LLMs) with a grasp detection method based on similarity matching with known templates. This approach enables accurate interpretation of intuitive human instructions and robust 6-DoF grasp generation by leveraging references from similar pre-existing instances. An example application of our proposed method is shown in Fig. \ref{img:1}. Given a previously unseen object and a natural human instruction, we first process the semantic information by associating the human instruction with an existing ontology and identifying the functional part of the target object in the given task. Then, we utilize the geometric information obtained from single-view observation and employ a sampling-based method to recognize the corresponding functional part, represented as a point cluster. Based on this recognition result, we query a similar model template containing prior segmentation and grasping knowledge as a reference. Finally, through a matching and planning process assisted by the template, a high-quality 6-DoF grasp pose is generated and executed to successfully accomplish the task. The effectiveness of our method is validated by extensive real-world experiments, notably demonstrating strong generalization to novel-category objects via the scalable integration of LLM-guided ontology and similarity-based planning. 

A recent study, ShapeGrasp \cite{ShapeGrasp}, introduces a similar TOG framework that also leverages LLMs and knowledge graphs for grasp inference. A common challenge we have addressed is the accurate segmentation of objects into functionally meaningful parts. While ShapeGrasp relies heavily on empirically defined thresholds for object geometric decomposition, our method achieves significantly greater robustness through a sampling-based strategy that avoids extensive manual parameter tuning. In addition, their approach is limited to top-down grasps positioned at object centroids, whereas our framework generates full 6-DoF grasps that are explicitly aware of contact stability. The idea of using similar models as templates originates from our earlier work \cite{Chen}, which addressed novel object pick-and-place by referencing an existing object database. However, its exclusive reliance on global object similarity is insufficient to address the nuanced demands of TOG scenarios, which constitutes the central focus of this research.

Our main contributions can be summarized as: 

\begin{figure}[t]
    \centering
    \includegraphics[width=0.95\linewidth]{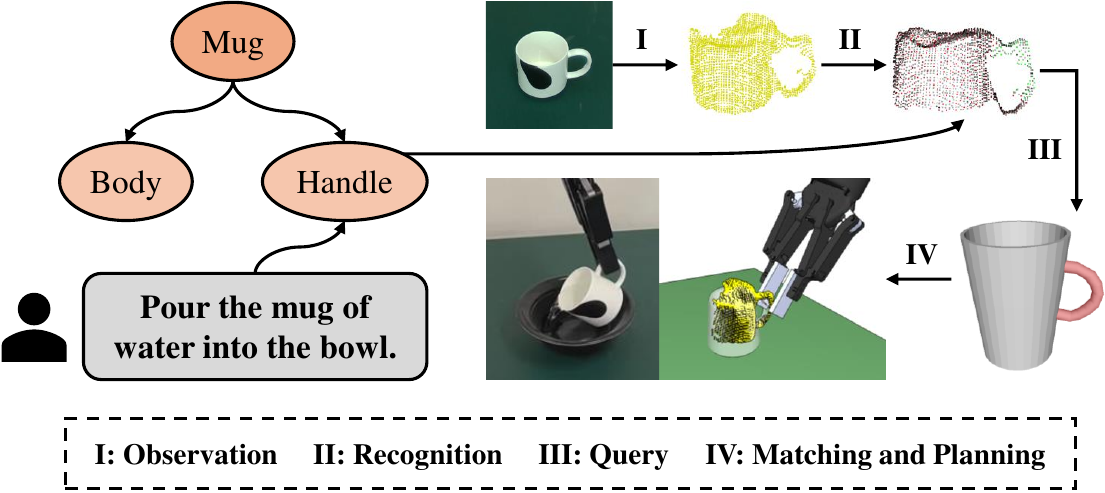}
    \caption{Demonstration of our method applied to a \textit{pouring} task with an unseen mug.}
    \label{img:1}
\end{figure}

\begin{itemize}

\item We propose an object-part-task ontology guided by LLMs that efficiently and accurately maps intuitive human instructions to the corresponding functional parts of target objects.

\item We introduce a geometry-based part recognition method that is robust to viewpoint variations by leveraging similar model templates as references for functional part matching and imitative grasp planning.

\item We optimize the quality of generated grasps from similar references through a combination of local-to-global point cloud registration and stability-aware positional adjustment.

\end{itemize}

\section{Related Works}
\subsection{Unseen Object Grasping}
Grasping research has explored strategies for handling unseen objects for decades, driven by the need to address the complexity of real-world environments, which are diverse, dynamic, and only partially understood. Among existing studies, model-based approaches using deep learning \cite{Pas,Mahler,Breyer,Jiang,Fang} or reinforcement learning \cite{Kalashnikov,Mandikal,Wang} dominate the field over model-free approaches based on geometric analysis \cite{Adjigble,Wu}. However, each approach has its strengths and limitations. Model-based methods exhibit high performance in specific grasping tasks through large-scale training, but are computationally expensive and sensitive to environmental changes. In contrast, model-free methods are cost-effective and robust across varying environments, but struggle to generalize to diverse object types.

To harness the benefits of model-free methods while enhancing generalizability, we investigate a promising direction based on similarity matching \cite{Chen,Goldfeder,Mitrevski}. The core idea of this approach is to leverage prior knowledge of similar known objects to guide the grasping of unknown targets. Inspired by \cite{Mitrevski}, we find that integrating similarity matching with ontological relationships can effectively extend the method's capabilities to tackle TOG beyond basic object handling.

\subsection{Task-Oriented Object Grasping}
High-level manipulation tasks require robots not only to grasp objects successfully, but also to grasp at appropriate positions to ensure both safety and functionality. To this end, TOG has been extensively studied, evolving from analytical approaches \cite{Dang,Song} to vision-based approaches \cite{Detry,Chu,Murali}, and more recently to vision-language-based approaches \cite{Tang,Wei,Zhang}. Here, we highlight a few recent notable works. CAGE \cite{Liu} is a representative affordance-aware grasping system considering both object and task constraints. GCNGrasp \cite{Murali} constructs a knowledge graph to train a Graph Convolutional Network, enabling the generalization of task-oriented grasps from predefined instances to novel concepts. OS-TOG \cite{Holomjova} and Robo-ABC \cite{Ju} propose matching frameworks that utilize database objects with labeled affordances to guide the grasping of novel objects. While conceptually similar to our approach, their methods are limited to 2D image matching, which restricts their ability to transfer 6-DoF grasping knowledge. Although all of these approaches effectively achieve generalized TOG, they do not incorporate language models and are therefore limited to simplistic task labels such as \textit{handover} or \textit{cut}. 

To address more complex human instructions, VLMs have increasingly been adopted for contextual understanding. LERF-TOGO \cite{Rashid} integrates CLIP embeddings \cite{Radford} with DINO features \cite{Caron} to train a multi-scale VLM capable of generating grasps for specific object parts based on language queries. GraspSplats \cite{Ji} introduces a Gaussian-based feature representation for real-time object motion tracking and enables dynamic language-guided manipulation. FoundationGrasp \cite{Tang2} leverages open-ended knowledge from foundation models (both LLMs and VLMs) to learn generalizable TOG skills. While these approaches successfully associate contextual language inputs with spatial motion outputs, they struggle to achieve high task success rates due to the complexity inherent in integrating vision and language within cross-domain learning frameworks. To cope with this issue, we draw inspiration from ontology-based grasping methods \cite{Wang2,Li2}, which defines multiple layers of knowledge representations (e.g., object, task, affordance) to support part-aware object manipulation. Rather than fusing all features into a unified model, we adopt a more modular approach: processing visual and verbal inputs separately and linking them through ontological reasoning over a predefined knowledge graph.

\subsection{LLMs in Task-Oriented Grasping}
LLMs are a crucial tool in our approach. In recent years, the emergence of high-performance LLMs such as \textit{ChatGPT} and \textit{Claude} has significantly advanced TOG strategies. For instance, GraspGPT \cite{Tang3} leverages LLMs to generate text descriptions for both objects and tasks, which are then encoded with object point clouds and grasp poses to select high-quality task-oriented grasps through a transformer decoder. ATLA \cite{Ren} utilizes LLMs to generate rich semantic knowledge, accelerating tool learning in diverse manipulation tasks. Grasp-Anything \cite{Vuong} employs LLMs to generate scene descriptions, which are integrated with image inputs using a diffusion model trained on a large dataset and applied to language-driven grasp detection. While these approaches commonly use LLMs to generate instructive text to enhance the learning process, we find that directly regressing grasp poses from text embeddings can lead to unstable performance due to the diversity of LLM descriptions and the uncertainty in real-world observations. Therefore, we propose a sequential strategy that first queries the target object parts using the strong interpretability of LLMs, and then focuses on object geometry for grasp planning. This separation effectively mitigates the complexity of handling linguistic and visual features simultaneously.

\section{Methods}
\subsection{Overview}
\begin{figure}[t]
    \centering
    \includegraphics[width=\linewidth]{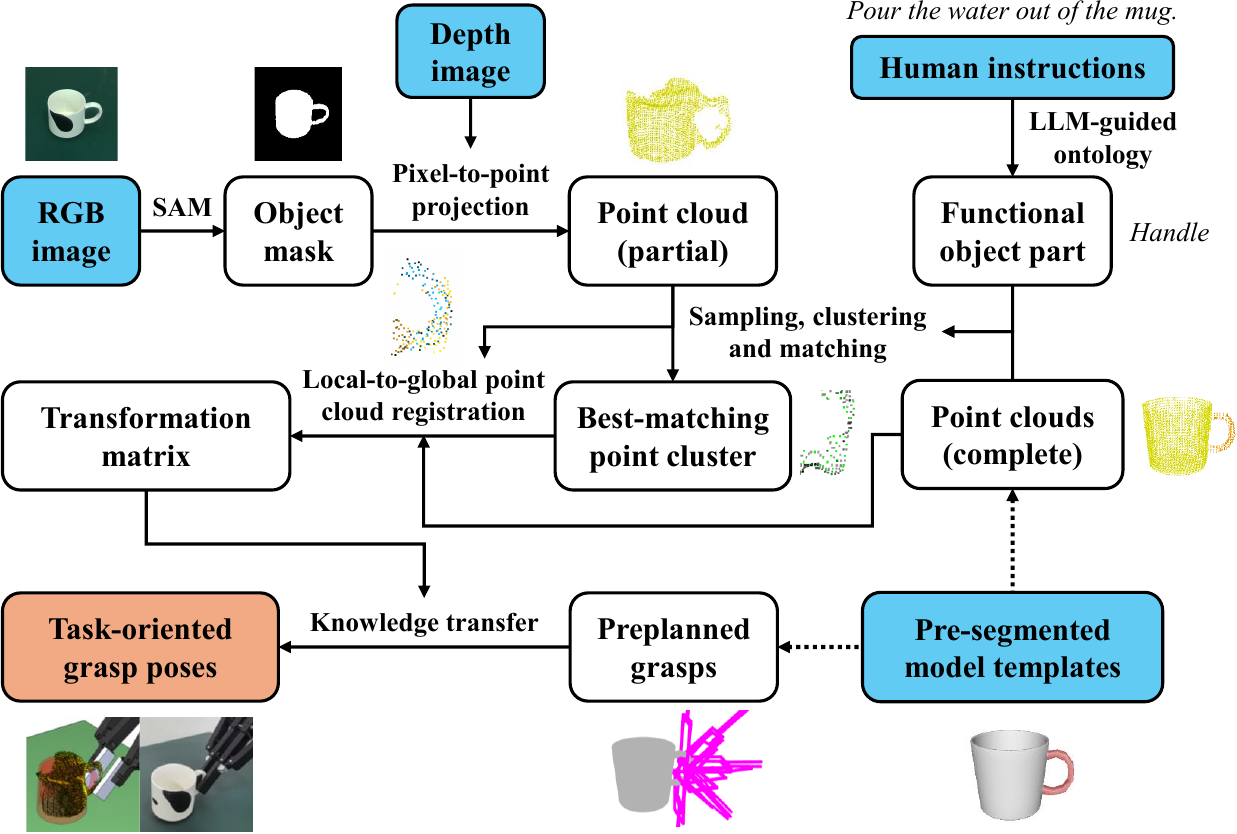}
    \caption{Proposed TOG system overview. Blue indicates inputs; orange indicates outputs.}
    \label{img:2}
\end{figure}

Our goal is to enable a robot to execute task-oriented grasps based on intuitive human instructions. Given an instruction, the robot uses an RGB-D camera to detect the target object, identify the relevant functional part, and plan feasible grasps that fulfill the intended task. The overall process consists of three core components: \textit{Human Instruction Interpretation}, \textit{Functional Part Recognition}, and \textit{Task-Oriented Grasp Planning}. A high-level overview of our TOG system is illustrated in Fig. \ref{img:2}.

On the left side, we process RGB-D images captured by a wrist-mounted camera through mask extraction and pixel-to-point projection, resulting in a partial point cloud of the target object. On the right side, we process human instructions using an LLM-guided ontology to identify the relevant functional part, which is used to guide the subsequent matching and planning stages. Meanwhile, we prepare a database of model templates, each segmented into multiple functional parts according to the defined ontology. In addition, each template is accompanied by a corresponding segmented point cloud and a set of preplanned robust grasps.

Based on the identified functional part and the pre-existing template point clouds, we adopt a sampling-clustering-matching strategy to recognize the corresponding part (a point cluster) within the observed object point cloud. We then perform local-to-global point cloud registration using both the full template point cloud and the selected point cluster to align the observed object with the best-matching model template, yielding a transformation matrix. This matrix is used to transfer preplanned grasping knowledge from the model template to the observed object. Finally, a dedicated optimization process generates robust task-oriented grasps tailored to the given instructions and object geometry for execution.

\subsection{Template Database Preparation}
The model template database used in our pipeline consists of three key components: pre-segmented object meshes, corresponding point clouds, and pre-planned grasps. While the point clouds and pre-planned grasps can be automatically generated using voxel grid downsampling and an existing antipodal grasp sampling method \cite{Wan}, the only manual step required in constructing the database is the part-aware segmentation of object meshes. However, this process is neither difficult nor time-consuming, due to the minimal requirement for precise labeling—the template serves merely as a reference rather than a direct target. Unlike pixel-level 2D image segmentation, our 3D mesh segmentation can be easily achieved by rotating the mesh to a suitable pose and using a simple rectangular selection box to roughly isolate the desired object part. This can be done using open-source platforms such as MeshLab \cite{Meshlab}. In practice, each mesh can be segmented in under 30 seconds.

It is also worth noting that our method does not heavily depend on the size or diversity of the template database. As demonstrated in Experiment \ref{gen}, even a small database with fewer than 10 model templates is sufficient to handle a wide variety of novel objects. This strong generalizability significantly reduces the preparatory workload required for database construction in real-world task scenarios, thereby keeping the data collection cost minimal.

\subsection{Human Instruction Interpretation}
In human-guided manipulation and human-robot collaboration tasks, accurately interpreting human instructions is essential for generating appropriate robot actions. However, the rich semantic knowledge embedded in contextual instructions is difficult to extract without a powerful inference model. Recent advances in LLMs have made this level of understanding feasible. In the context of TOG, a key challenge is determining \textit{which part of the object the robot should grasp}—the central problem addressed in this work. To associate human instructions with relevant object parts, we propose an LLM-guided object-part-task ontology.

\begin{figure}[t]
    \centering
    \includegraphics[width=0.85\linewidth]{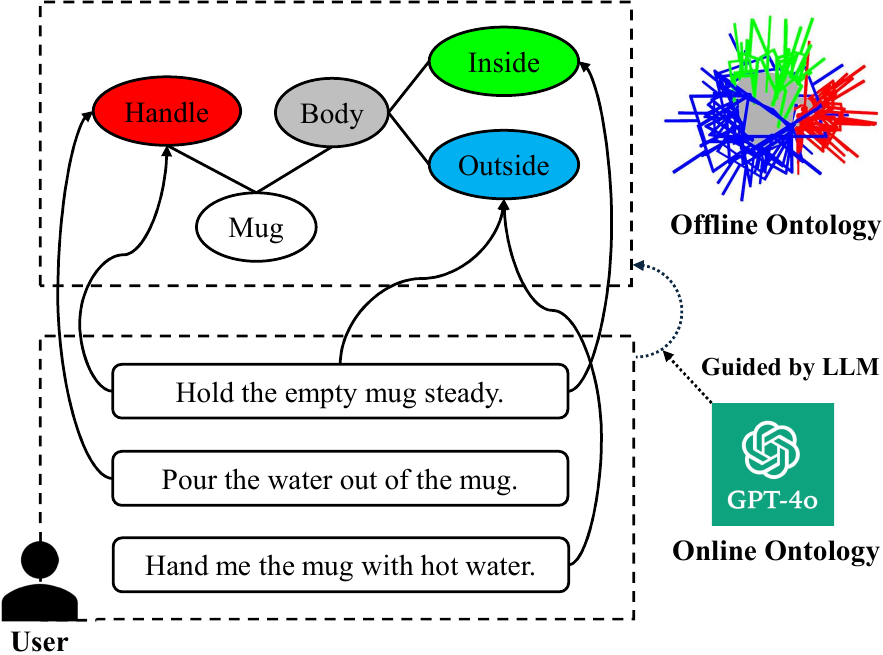}
    \caption{LLM-guided object-part-task ontology with online and offline components.}
    \label{img:3}
\end{figure}

As shown in Fig. \ref{img:3}, the proposed ontology consists of two components: offline and online. The offline ontology predefines object classes with multi-layer part hierarchies. For instance, a \textit{mug} is divided into \textit{handle} and \textit{body}, while the \textit{body} can be further subdivided into \textit{inside} and \textit{outside}. Different object parts correspond to different grasping strategies, which are flexibly selected according to task requirements. The online ontology handles user-provided instructions, which are not previously known. To establish connections between task instructions and relevant functional parts, we leverage the latest LLM, GPT-4o \cite{OpenAI}, to bridge the online and offline components. The prompt used for this association is originally structured as follows:

\textit{
Given the following ontology ... A robot is given the following commands ... Question: Which part(s) of the object should the robot grasp?
}

Despite the strong interpretability of LLMs, we observe that their answers can vary across trials, lacking accuracy and consistency. To address this, we first adopt the prompt optimization method from \cite{Yang}, which uses LLMs themselves as prompt optimizers. Implementation details are provided in \ref{A}. While the optimized prompts incorporating task constraints improve answer correctness, they still fall short of human-level reliability. Through a large number of trials, we identify two prompt design principles that significantly enhance LLM performance: 1) Step-by-step reasoning; 2) Using an answer template. In practice, we append a step-by-step reasoning template to the end of each prompt, following this structure:

\textit{
The command is ...
Step 1: Identify the type of task ...
Step 2: Apply task constraints ...
Analyzing the object parts ...
Best choice for the robot ...
\textbf{Conclusion: The robot should grasp ...}
}

The functional part name can be extracted from the \textbf{Conclusion} line. For further evaluation, we apply this structured prompt to the natural language annotations provided in the DROID dataset \cite{Khazatsky}. We select 20 task instructions that are suitable for TOG scenarios, such as
\textit{"Pour the contents of the orange cup into the pot"} and 
\textit{"Use the sharpie to draw on the board"}.
For each target object in each task, we manually define a corresponding offline ontology and assign ground-truth labels for functional part names. To assess repeatability, each instruction is tested 3 times using different agents. Across a total of 60 trials, the LLM responses achieve 100\% accuracy, demonstrating the reliability and stability of the proposed ontological reasoning approach.

\subsection{Functional Part Recognition}\label{rec}
\begin{figure}[t]
    \centering
    \includegraphics[width=0.9\linewidth]{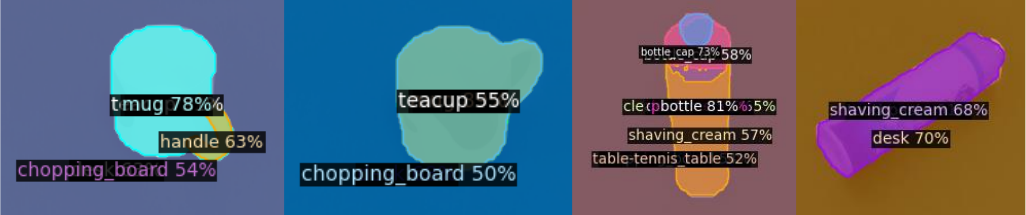}
    \caption{Sensitivity of model-based part segmentation to viewpoint changes (colors in the figure are randomly assigned during segmentation and do not indicate part identity).}
    \label{img:4}
\end{figure}

To map the semantic identification of functional object parts into geometric space for subsequent planning, we need to extract part-level features from visual inputs. A common approach is to train a vision model for object part segmentation. However, we observe that even state-of-the-art segmentation models \cite{Sun} suffer from viewpoint sensitivity, where segmentation results become inconsistent under varying viewpoints. As shown in Fig. \ref{img:4}, components such as the \textit{mug handle} and the \textit{bottle cap} are not reliably identified when the objects are placed in different poses. To address this limitation, we propose a new strategy based on template-assisted geometric analysis.

\begin{figure}[t]
    \centering
    \includegraphics[width=\linewidth]{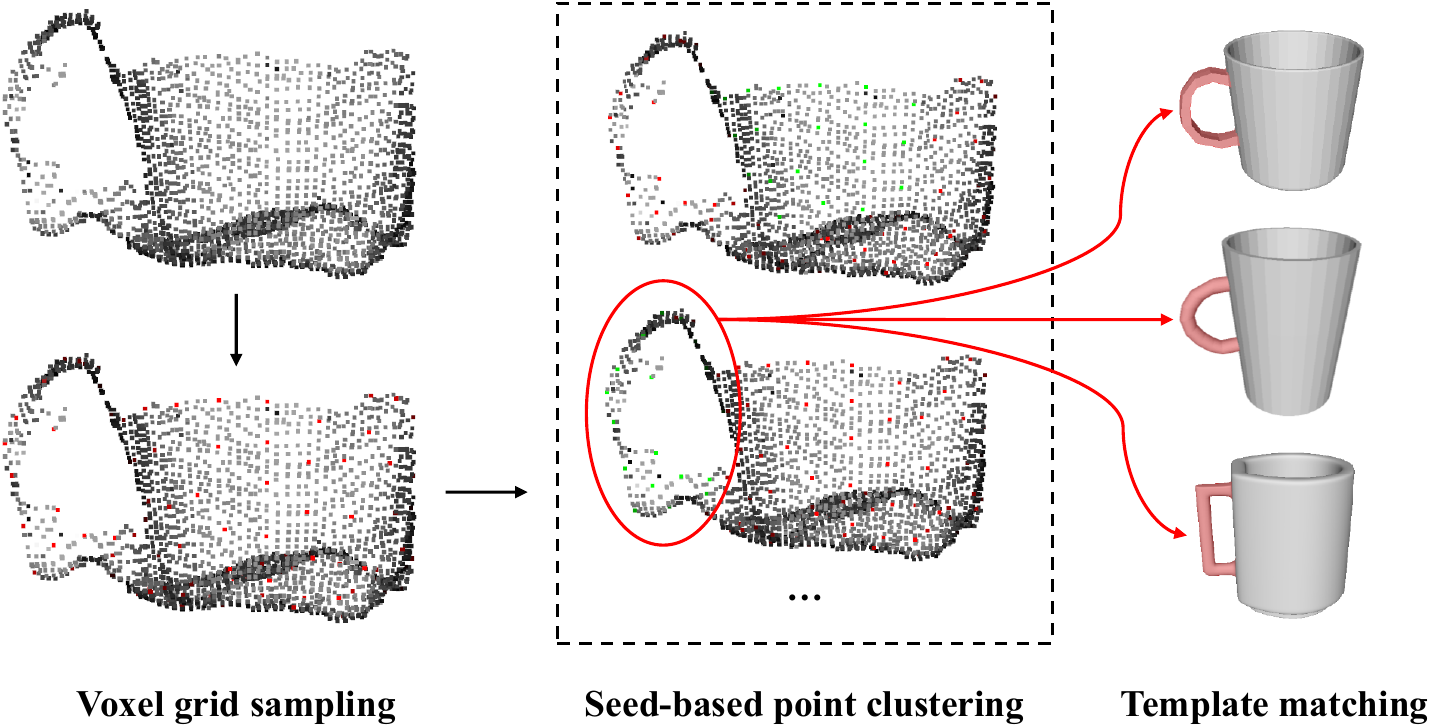}
    \caption{Template-assisted object part recognition using a three-step strategy.}
    \label{img:5}
\end{figure}

Given the RGB input containing the target object, we first apply a powerful class-agnostic segmentation method that is robust to viewpoint variations, SAM \cite{Kirillov}, to obtain an object mask indicating the object's planar location. Using the corresponding depth input, we perform pixel-to-point projection to map the 2D mask into a 3D point cloud, capturing the object's spatial geometry. However, due to partial observation, the resulting object point cloud often contains large unseen areas, making it difficult to perform reliable part segmentation with incomplete geometric features. To address this, we incorporate pre-segmented model templates associated with the predefined ontology and use them as references to guide part segmentation. This template-guided process consists of three main steps: sampling, clustering and matching, as illustrated in Fig. \ref{img:5}. We begin by converting all model templates from mesh representations to point clouds using a voxel grid filter with an appropriate leaf size (5 mm in our task). Then, in the sampling step, we apply the same voxel filter to downsample the observed object point cloud, ensuring uniform point density. Next, we take each sampled point as a seed and search for its $k$-nearest neighbors to form a local point cluster. The number of neighbors $k$ is determined using the following equation:
\begin{equation}
    \label{eqn:1}
    k=\frac{N(o_{all}) N(m_{part})}{N(m_{all})} 
\end{equation}
where $N(\cdot)$ denotes the number of points in a point cloud. $o_{all}$ and $m_{all}$ represent the entire point clouds of the observed object and the model template, respectively. $m_{part}$ corresponds to the point cloud of the functional part within the template. Finally, the sampled point clusters are matched against the pre-segmented model templates to identify the best-matching cluster, the one that most closely resembles the target functional part.

However, the matching process is non-trivial. As illustrated in Fig. \ref{img:5}, point clouds captured by consumer-grade cameras often suffer from low precision due to sensing noise. To ensure the accuracy, efficiency, and stability of the matching process under such conditions, we introduce a multi-metric similarity evaluation method that considers both local and global geometric similarity. Letting a sampled point cluster be denoted as $o_{part}$, we first apply Principal Component Analysis (PCA) to evaluate the similarity in point distribution between $o_{part}$ and $m_{part}$, using the following metric:
\begin{equation}
    \label{eqn:2}
    d_{pca}=\left \| \frac{\sigma_o}{\left | \sigma_o \right | } - \frac{\sigma_m}{\left | \sigma_m \right | } \right \|_2 
\end{equation}
where $d_{pca}$ represents the PCA-based distributional difference between $o_{part}$ and $m_{part}$. $\sigma_o$ and $\sigma_m$ denote the 3-dimensional PCA singular value vectors of $o_{part}$ and $m_{part}$, respectively, normalized by their Euclidean norms. This evaluation allows for a quantitative comparison between point distributions. However, PCA alone is insufficient to fully capture shape similarity. An intuitive example is that a cube and a sphere may yield similar PCA results due to geometric symmetry, despite having fundamentally different shapes.

\begin{figure}[t]
    \centering
    \includegraphics[width=\linewidth]{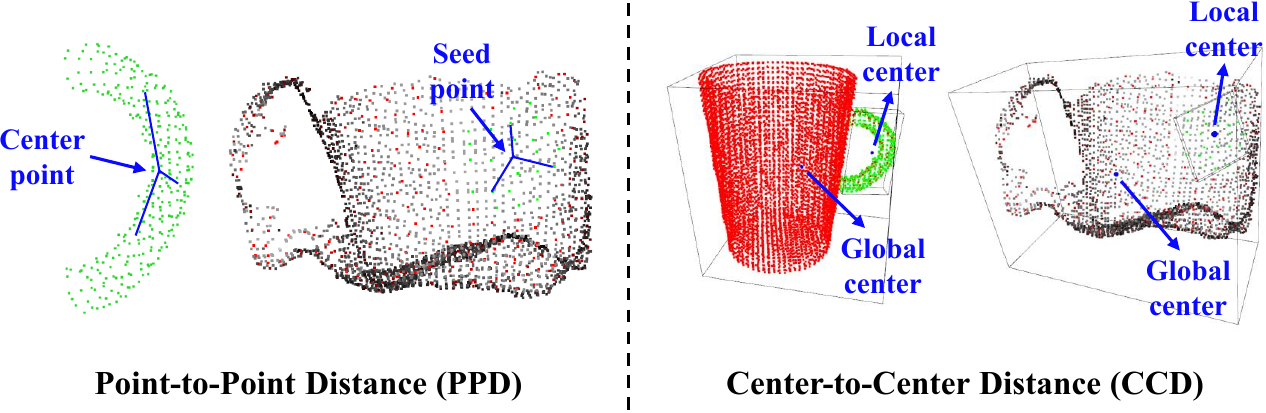}
    \caption{Two distance metrics for evaluating similarity between point clusters.}
    \label{img:6}
\end{figure}

To address this issue, we introduce a second evaluation metric: Point-to-Point Distance (PPD). As illustrated on the left side of Fig. \ref{img:6}, we use the center point (for $m_{part}$) and the seed point (for $o_{part}$) to calculate the PPD. The center point is identified by first computing the center of the bounding box of $m_{part}$, and then finding its nearest neighbor within $m_{part}$. Based on this point, we calculate the standard deviation of its distances to all other points in $m_{part}$, denoted as $D(s_m)$. For the seed point sampled in $o_{part}$, we similarly compute the standard deviation of its distances to all other points in the local cluster, denoted as $D(s_o)$. The PPD is then evaluated as:
\begin{equation}
    \label{eqn:3}
    d_{ppd}=\left | \frac{D(s_o)}{\max(\left |s_o-\overline{s_o}\right |)} - \frac{D(s_m)}{\max(\left |s_m-\overline{s_m}\right |)} \right | 
\end{equation}
where $\max(\cdot)$ denotes the maximum deviation from the mean occurring during the point distance calculation. The resulting value, $d_{ppd}$, reflects the difference in point dispersion between $m_{part}$ and $o_{part}$, offering an additional cue for distinguishing functional parts with different spatial shapes. However, this metric is non-directional and does not capture the full 3D geometric structure, making it unsuitable for use in isolation.

While the combination of $d_{pca}$ and $d_{ppd}$ performs well for evaluating similarity between point clusters, they focus primarily on local geometry and can become less reliable when visual features are sparse or noisy. To address this limitation, we incorporate a third metric: Center-to-Center Distance (CCD). As illustrated on the right side of Fig. \ref{img:6}, we obtain the bounding box centers of both the entire point clouds $m_{all}, o_{all}$ and the part point clouds $m_{part}, o_{part}$, denoted as global center and local center, respectively. The CCD is then evaluated as:
\begin{equation}
    \label{eqn:4}
    d_{ccd}=\left | \frac{c_o}{b_o} - \frac{c_m}{b_m} \right | 
\end{equation}
where $c_o$ and $c_m$ represent the distances between the local and global centers within the observed object and the model template, respectively. $b_o$ and $b_m$ denote the maximum possible distances, defined as half the diagonal length of the bounding boxes of $o_{all}$ and $m_{all}$, respectively. The resulting value, $d_{ccd}$, reflects the difference in the relative global positions of $m_{part}$ and $o_{part}$ within their respective entire objects, thereby enhancing functional part recognition especially when local visual features are sparse or noisy.

Combining the three metrics, we define the final evaluation function as:
\begin{equation}
    \label{eqn:5}
    d= d_{pca} + d_{ppd} + d_{ccd}
\end{equation}
where $d$ represents the overall difference between a template functional part and a sampled point cluster within the observed object. Smaller values of $d$ indicate higher similarity. Note that in Equations \ref{eqn:2}-\ref{eqn:4}, all metrics have been normalized using appropriate boundary values, allowing them to be combined directly through addition. The final output of the matching process is the point cluster $o^*_{part}$ that achieves the lowest average $d$ across matches with $m_{part}$ from all available templates.

\subsection{Task-Oriented Grasp Planning}
After recognizing the functional part represented by $o^*_{part}$, the next step is to plan robust grasps within its region to achieve the goal of TOG. However, due to the sparse and noisy point features in $o^*_{part}$, directly generating reliable grasps from the point cloud is extremely challenging. To overcome this, we leverage the matched model template again to transfer grasping knowledge from known objects to unknown targets. For each model template, we use a mesh-segmentation-based approach \cite{Wan} to preplan over 50 grasps on each subdivided part based on the offline ontology. For a given task, only the preplanned grasps associated with the relevant functional part are utilized. To transfer this grasping knowledge from the template to the object, we first need to obtain the transformation matrix between the point clouds $o_{all}$ and $m_{all}$ to determine their relative pose. However, the incomplete point cloud from partial observation poses a challenge for accurate alignment with the complete template point cloud using traditional registration methods like ICP \cite{ICP}. To address this, we propose a novel strategy called local-to-global registration, which aligns the point clouds step by step, from the functional part to the entire object, thereby enhancing alignment accuracy.

\begin{figure}[t]
    \centering
    \includegraphics[width=\linewidth]{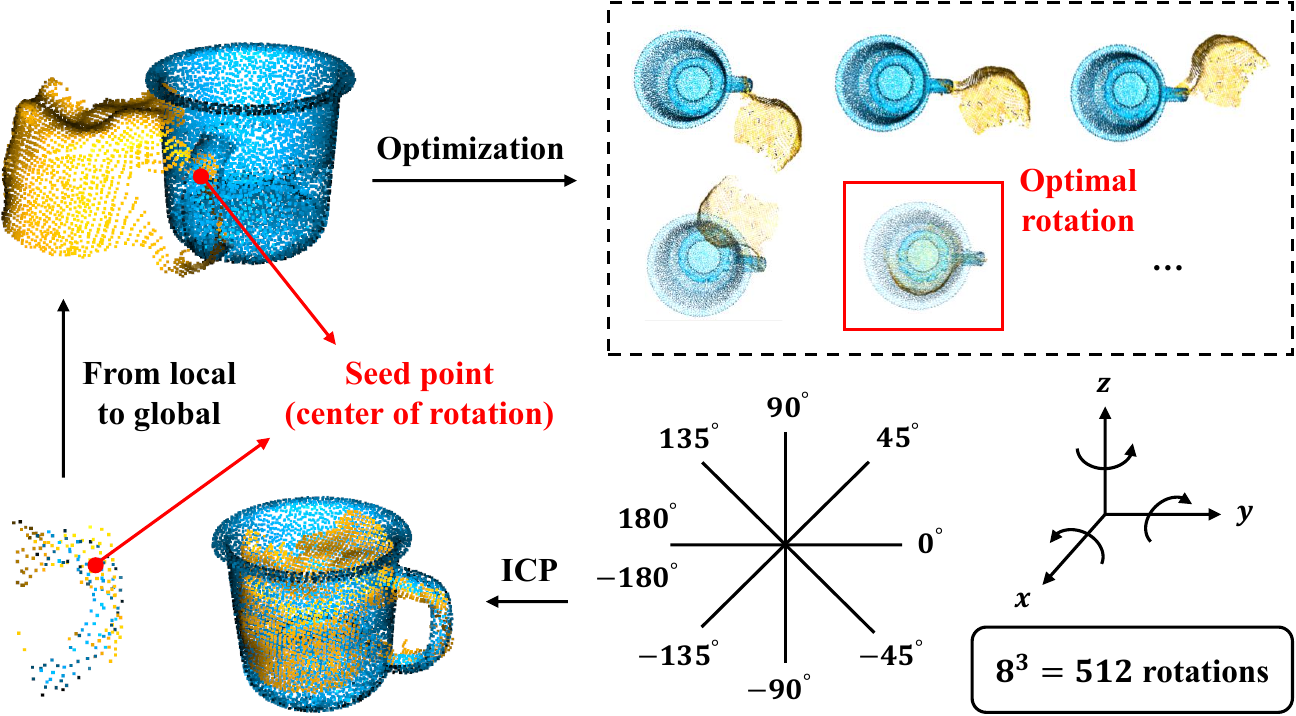}
    \caption{Local-to-global point cloud registration through rotation optimization.}
    \label{img:7}
\end{figure}

As illustrated in Fig. \ref{img:7}, we begin by matching $o^*_{part}$ with $m_{part}$ using a combined registration algorithm, RANSAC + ICP, implemented in Open3D\footnote{http://open3d.org/docs/release/tutorial/pipelines}. This process is repeated iteratively until a sufficient number of point correspondences (over half of the points in $o^*_{part}$) are identified, indicating that the two point clouds are well-aligned. Next, we apply the resulting transformation matrix, denoted as $T_{loc}$, to initially align $o_{all}$ with $m_{all}$. At this stage, although the functional parts are aligned, the entire objects may still show some misalignment due to rotational errors in the local registration. Directly applying ICP at this point often fails to improve the result. To overcome this, we introduce an optimization method that rotates $o_{all}$ around the seed point of $o^*_{part}$ to find an optimal rotation, denoted as $T_{opt}$, which minimizes the distance between $o_{all}$ and $m_{all}$. The rotation space is discretized by sampling angles from $-180^\circ$ to $180^\circ$ in $45^\circ$ increments along each of the three axes, resulting in $8^3=512$ total rotation attempts. This optimization process does not take long with parallel computing. Once the optimal rotation is identified, we apply the ICP algorithm to further refine the alignment between $o_{all}$ and $m_{all}$, ensuring precise registration. Assuming the resulting transformation matrix from ICP is $T_{icp}$, the final transformation from $o_{all}$ to $m_{all}$ is then calculated as $T=T_{loc}T_{opt}T_{icp}$. 

While this local-to-global registration is performed between the target object and all relevant model templates, we use only the best-matching template, which achieves the highest fitness score in the final ICP process, as the reference for grasp planning. Assuming the set of preplanned grasps on the best-matching template is $G_m$, and the camera pose relative to the robot base (the origin of the world coordinate system) is $T_0$, we can generate a corresponding set of imitative grasps on the target object as $G_o=T_0T^{-1}G_m$. To ensure executable grasps for real-world tasks, we perform IK computation and collision detection within a simulation environment\footnote{https://github.com/wanweiwei07/wrs}, filtering out infeasible grasps that are either unreachable or result in collisions. For selecting reliable grasps positioned on the functional part, we prioritize those located within the region of $o^*_{part}$ by generating a cube collision model $M_{cube}$ within the gripper closure area and a point cloud collision model $M_{pcd}$ corresponding to $o^*_{part}$. When $M_{cube}\cap M_{pcd}\ne \emptyset$, the grasp is considered correctly positioned and selected for execution.

\begin{figure}[t]
    \centering
    \includegraphics[width=0.9\linewidth]{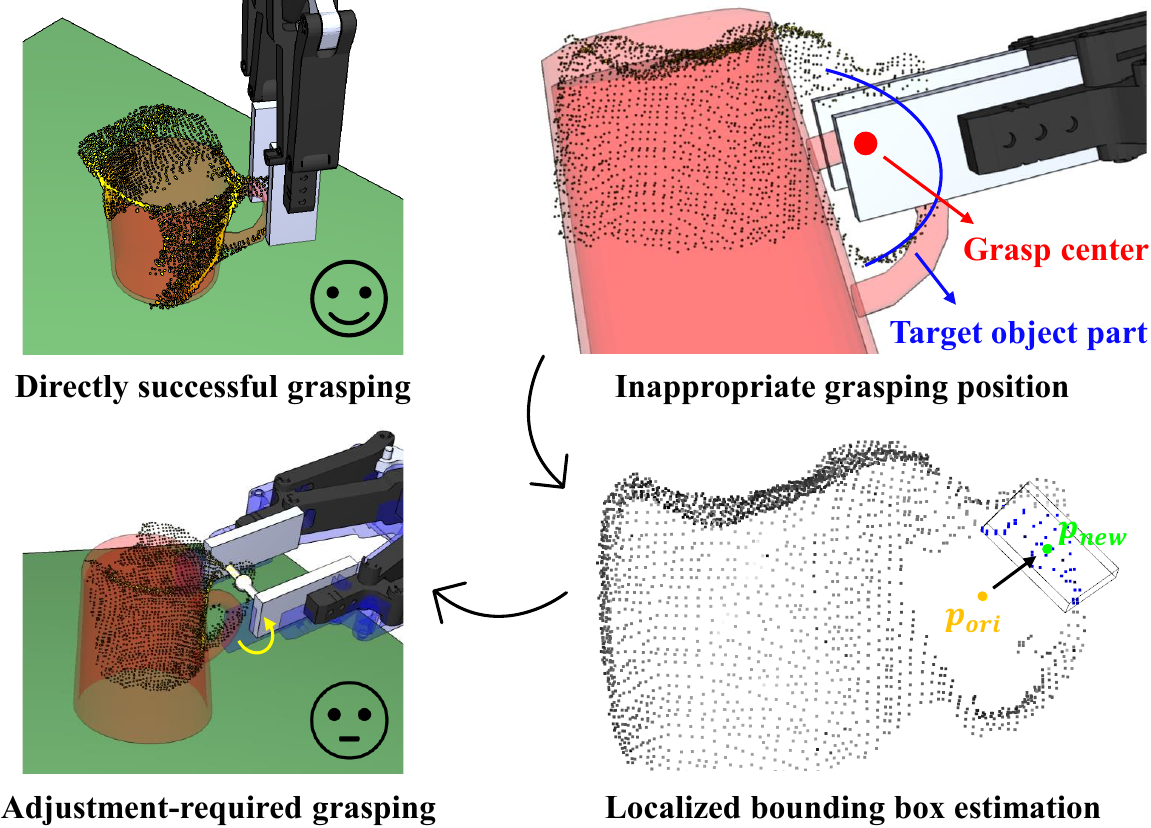}
    \caption{Adjustment of suboptimal grasps via localized bounding box estimation.}
    \label{img:8}
\end{figure}

However, we observe that the transferred imitative grasps are not always highly stable due to inherent differences between the model template and the target object. As shown in Fig. \ref{img:8}, when the two point clouds align perfectly, the generated grasps are typically executable without failure. However, when small deviations occur, misalignment between the grasp center and the target object part may introduce potential risks, such as: 1) Collision between the grasp pose and other parts of the object; 2) Suboptimal contact between the gripper and the object, which may reduce grasp stability. To detect such conditions, we generate a stick collision model $M_{stick}$ connecting the two ends of the gripper. When $M_{stick}\cap M_{pcd}= \emptyset$, the grasp center is considered misaligned with the target object part and requires adjustment. Assuming the original grasp center is $p_{ori}$, we search for its nearest neighbor point in $o^*_{part}$, denoted as $p_{neg}$. We then search for the $k$-nearest neighbors of $p_{neg}$ and estimate a localized bounding box (LBB) based on these points. Here, the value of $k$ is approximately set to half the number of points in $o^*_{part}$. The center of the LBB is adopted as the new grasp center, denoted as $p_{new}$. Finally, the original grasp pose is translated along the vector $\overrightarrow{p_{ori} p_{new}}$ to let the grasp center align with the target object part, thereby improving stability.

\section{Experiments}
\subsection{Experimental Setup}
To validate the effectiveness of our proposed method, we conduct the following TOG experiments: 1) a benchmark study on grasping various types of unseen objects under different human instructions; 2) an ablation study to evaluate the local-to-global registration and the stability-aware grasp adjustment; and 3) an investigation into the generalization to novel-category objects that are not part of the existing ontology. All experiments are performed in the real world, using a UR5e robot arm equipped with a Robotiq 2F-140 adaptive gripper and a wrist-mounted RealSense D435 RGB-D camera. The experimental objects are placed on a fixed platform within the robot's reachable workspace, and are observed by the wrist camera from a diagonal downward view. All computations are carried out on a computer equipped with a Ryzen 7 5800H CPU and a GeForce RTX 3060 GPU.

\subsection{Task-Oriented Grasping of Unseen Objects}
\begin{figure}[t]
    \centering
    \includegraphics[width=\linewidth]{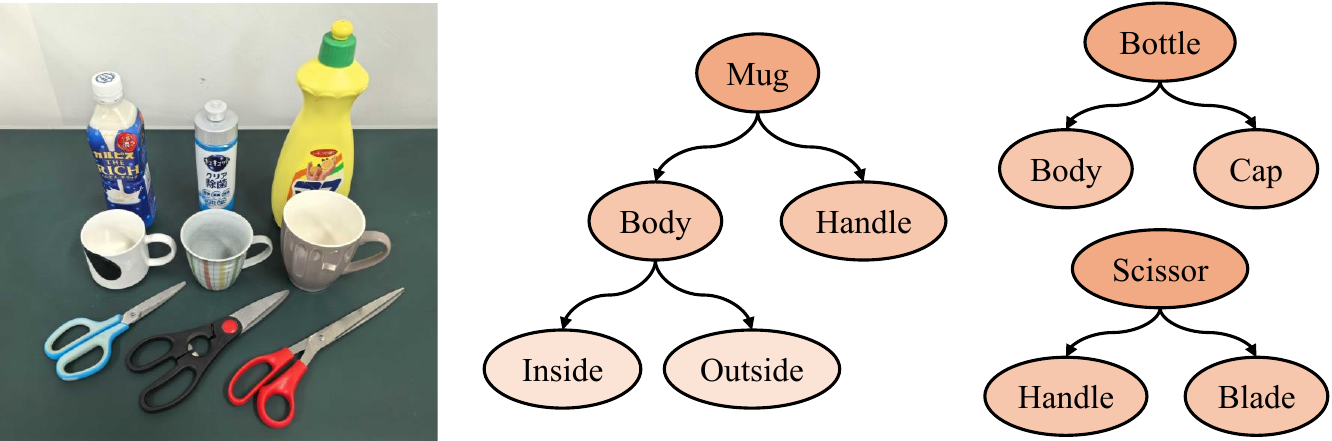}
    \caption{Test objects used in TOG experiments with predefined ontological knowledge.}
    \label{img:9}
\end{figure}

For the first TOG experiment, we select three types of divisible objects with functional parts: \textit{mug}, \textit{bottle}, and \textit{scissor}. Each category includes three previously unseen items with varying shapes and sizes, and their predefined ontological knowledge is stored in a database, as shown in Fig. \ref{img:9}. Meanwhile, we incorporate a set of model templates extracted from open-source libraries such as GrabCAD\footnote{https://grabcad.com/library} into the database. Each template is pre-segmented according to the predefined ontology, and is accompanied by downsampled point clouds and preplanned grasping knowledge for each functional part.

\subsubsection{Evaluation of Part Recognition Accuracy}\label{exp}
We begin by evaluating our geometry-based part recognition method through a comparison with a state-of-the-art VLM-based approach, VLPart \cite{Sun}. We select three salient functional parts: \textit{mug handle}, \textit{bottle cap}, and \textit{scissor handle}, as target regions to facilitate a clear assessment of recognition performance. Each object is placed in 5 randomized poses with their functional parts visible to the camera. For the VLPart trials, we capture RGB images and apply a pre-trained segmentation model based on Cascaded Swin Transformers, using a custom vocabulary defined by the name of the target functional part. For our method, we capture object point clouds and employ the proposed template matching approach to identify the target functional part as a point cluster. To ensure a fair evaluation, we map this point cluster to a mask region in the RGB image and represent the functional part with a minimal 2D bounding box, in the same manner as VLPart. Ground truth labels are manually assigned by fitting bounding boxes to the exact target regions. A recognition is considered accurate if the Intersection over Union (IoU) exceeds 0.5. The final part recognition accuracy (PRA) is calculated~as:
\begin{equation*}
    \mathrm{PRA= \frac{Number\;of\;correctly\;recognized\;parts}{Number\;of\;total\;recognition\;attempts}} 
\end{equation*}

\begin{table}[t]
\small
\renewcommand\arraystretch{1.5}
\setlength\tabcolsep{10pt}
\centering
\begin{tabular}{c | c c c | c}
\toprule
Object Class & Mug & Bottle & Scissor & \multirow{2}{*}{Average PRA} \\ 
Functional part & Handle & Cap & Handle \\
\hline
VLPart & 12/15 & 11/15 & 6/15 & 29/45 (64.4\%) \\
Our method & 12/15 & 13/15 & 15/15 & \textbf{40/45 (88.9\%)} \\
\bottomrule
\end{tabular}
\caption{Evaluation of PRA via a Baseline Study Against a VLM-Based Approach}\label{tab:1}
\end{table}

\begin{figure}[t]
    \centering
    \includegraphics[width=0.95\linewidth]{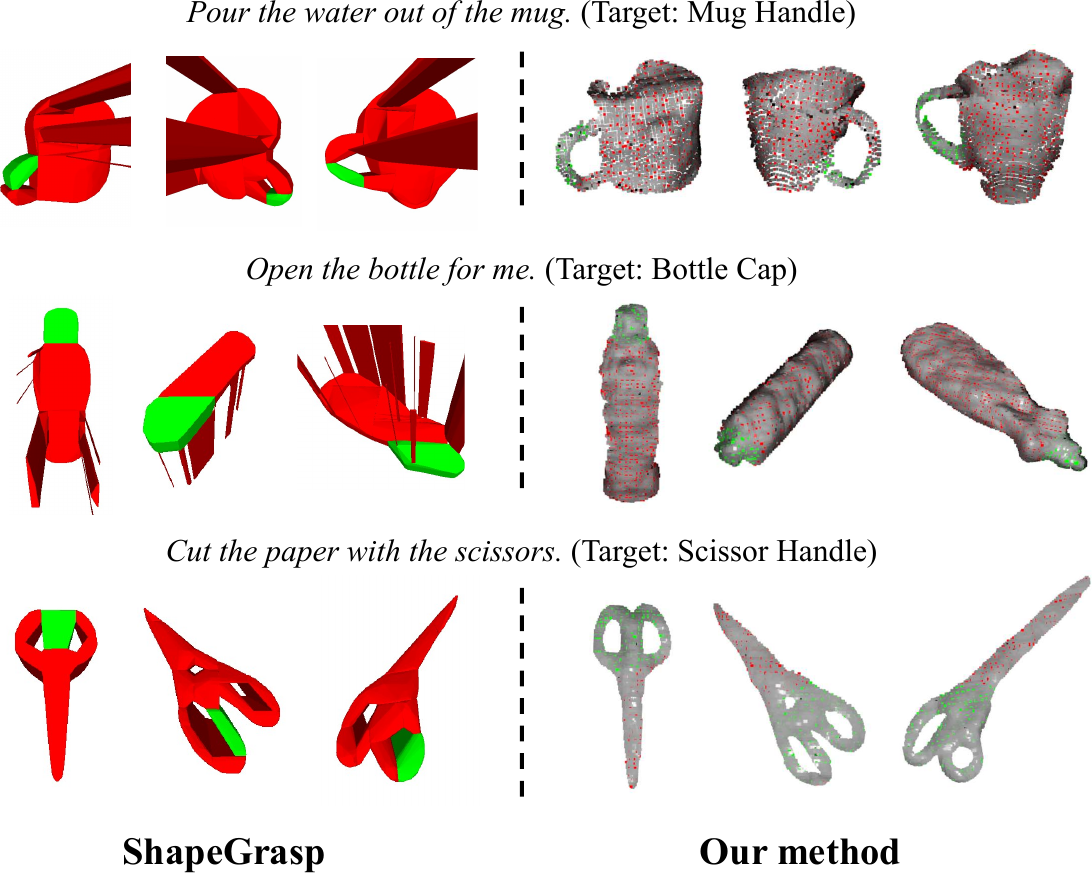}
    \caption{Task-oriented functional part recognition: ShapeGrasp vs. Our method.}
    \label{img:11}
\end{figure}

The experimental results are presented in Table \ref{tab:1}. Our method significantly outperforms VLPart in recognizing various functional parts across different object categories. Notably, in the case of the \textit{scissor handle}, VLPart exhibits frequent failures with the red scissor featuring an irregular handle, whereas our method maintains consistent performance. This performance gap is primarily attributed to the viewpoint sensitivity of model-based approaches, as discussed in Section \ref{rec}. In contrast, our method mitigates this issue by leveraging 3D geometric features, which remain relatively invariant under changes in viewpoint. Fig. \ref{img:11} illustrates the robustness of our method in recognizing various functional object parts from different viewpoints.

\begin{figure}[t]
    \centering
    \includegraphics[width=0.8\linewidth]{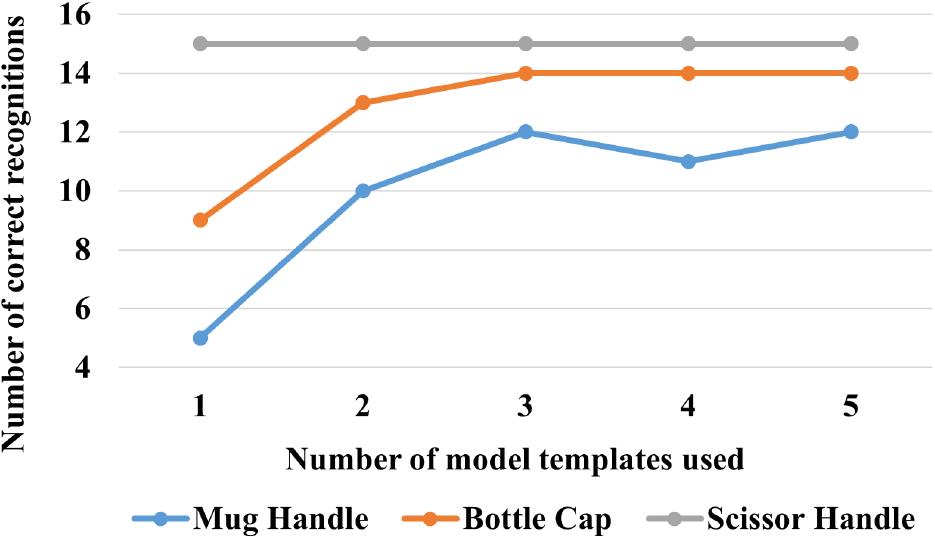}
    \caption{Exploring the impact of template quantity on part recognition performance.}
    \label{img:10}
\end{figure}

Additionally, we investigate the impact of model template quantity on part recognition performance. For each object class, we prepare 5 model templates that vary in shape and size. During evaluation, we progressively increase the number of templates used for similarity matching from a single template up to all five. As in the previous experiment, we conduct 5 trials per object under each condition and record the number of correct recognitions. As shown in Fig. \ref{img:10}, both the \textit{mug cap} and \textit{bottle cap} reach optimal performance when the number of templates increases to 3, with no further improvement observed beyond that point. This suggests that using multiple templates helps average out matching errors and enhance recognition accuracy; however, excessive template inclusion yields no additional benefit. On the other hand, object parts with distinctive geometric features, such as the \textit{scissor handle}, demonstrate high recognition accuracy even with a minimal number of templates. Based on these findings, we standardize the use of 3 model templates per object class in our TOG framework.

\subsubsection{Evaluation of Grasp Selection Accuracy}
Next, we evaluate our grasp selection method against ShapeGrasp \cite{ShapeGrasp}, a recent task-guided approach that also utilizes LLM reasoning but employs a decomposition graph for functional part recognition, and GraspGPT \cite{Tang3}, a recent learning-based method that relies on neural network inference. Due to GraspGPT's inability to generate real-time grasps and ShapeGrasp's restriction to top-down grasping, this experiment focuses specifically on selecting appropriate grasping positions based on human instructions, rather than on grasp stability. To achieve this, we first employ our similarity-based planning method to pre-generate a set of feasible grasp candidates densely distributed across all object parts, and then apply different baselines for grasp pose selection. It should be noted that other grasp planning approaches can serve as alternatives; however, we find our method to be the most efficient in generating part-level dense grasps. Additionally, the grasp generation process is separated from functional part reasoning and recognition, ensuring no bias in performance comparisons. For all baselines, the inputs consist of object point clouds, pre-generated grasp candidates, and a natural language instruction. The output is a single optimal grasp selected either through neural inference (GraspGPT), graph searching (ShapeGrasp), or ontological reasoning (our method). For both ShapeGrasp and our method, the selected grasp is the first found candidate located within the identified functional part region. For evaluation, we design the following task instructions as language inputs, each targeting a specific functional part of the selected objects:

1) \textit{Pour the water out of the mug.} (Mug $\to$ Handle)

2) \textit{Hold the coffee-filled mug steady.} (Mug $\to$ Body $\to$ Outside)

3) \textit{Shake the bottle before I drink it.} (Bottle $\to$ Body)

4) \textit{Open the bottle for me.} (Bottle $\to$ Cap)

5) \textit{Cut the paper with the scissors.} (Scissor $\to$ Handle)

6) \textit{Hand the scissors to me.} (Scissor $\to$ Blade)

Instructions 1, 3, and 5 correspond to robot manipulation tasks, while Instructions 2, 4, and 6 involve human-robot interaction. Each instruction is tested on 5 object point clouds captured from different viewpoints during the recognition experiments. A selection is considered accurate if the chosen grasp lies within the region of the target functional part. The final grasp selection accuracy (GSA) is calculated as:
\begin{equation*}
    \mathrm{GSA= \frac{Number\;of\;correctly\;selected\;grasps}{Number\;of\;total\;selection\;attempts}} 
\end{equation*}

\begin{table}[t]
\small
\renewcommand\arraystretch{1.5}
\setlength\tabcolsep{4pt}
\centering
\begin{tabular}{c | c c | c c | c c | c}
\toprule
Object Class & \multicolumn{2}{c|}{Mug} & \multicolumn{2}{c|}{Bottle} & \multicolumn{2}{c|}{Scissor} & \multirow{2}{*}{Average GSA} \\ 
Instruction No. & 1 & 2 & 3 & 4 & 5 & 6 \\
\hline
GraspGPT & 10/15 & 9/15 & 14/15 & 3/15 & 12/15 & 5/15 & 53/90 (58.9\%) \\
ShapeGrasp & 12/15 & 10/15 & 15/15 & 8/15 & 13/15 & 8/15 & 66/90 (73.3\%) \\
Our method & 11/15 & 14/15 & 15/15 & 11/15 & 15/15 & 13/15 & \textbf{79/90 (87.8\%)} \\
\bottomrule
\end{tabular}
\caption{Evaluation of GSA Against Two State-of-the-Art Task-Guided Approaches}\label{tab:2}
\end{table}

\begin{figure}[t]
    \centering
    \includegraphics[width=\linewidth]{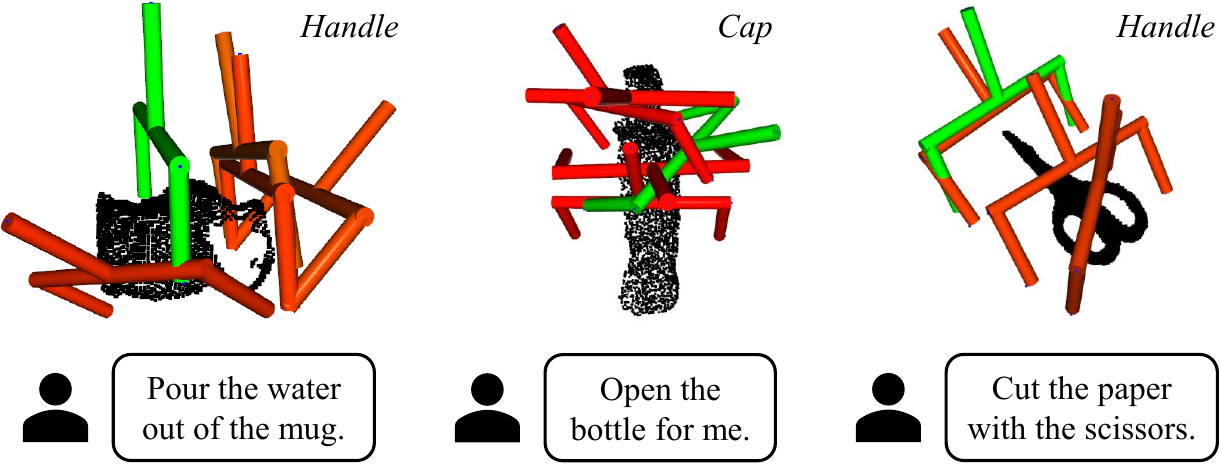}
    \caption{Failure cases of GraspGPT in language-guided grasp selection. Selected grasps are highlighted in green, while the ground truth labels are shown in the top-right corner.}
    \label{img:12}
\end{figure}

The experimental results are presented in Table \ref{tab:2}. Our method significantly outperforms both GraspGPT and ShapeGrasp in language-guided grasp selection. Notably, we observe substantial performance gaps between the methods for Instructions 4 and 6. For GraspGPT, we attribute this gap to two main reasons: 1) The \textit{bottle} category is absent from the training dataset, and the \textit{cap}, being a small and distinctive part of the \textit{bottle}, is likely overlooked during grasp selection. 2) The \textit{scissor} requires careful handling in a handover task, whereas GraspGPT does not incorporate safety constraints, such as \textit{the robot should handle dangerous parts instead of the human}. Example failures are shown in Fig. \ref{img:12}. For ShapeGrasp, we observe significant deformation on the backside of the objects (the area not visible to the camera) when reconstructing 3D meshes from 2D masks, as shown in Fig. \ref{img:11}. This deformation introduces a risk of misalignment between the predicted grasp position and the actual part position. Additionally, the mesh decomposition in ShapeGrasp is highly sensitive to parameter variations, which sometimes leads to the recognized functional part being either smaller than the actual region (e.g., \textit{mug handle} and \textit{scissor handle}) or larger than the actual region (e.g., \textit{bottle cap}). In contrast, our method requires no parameter tuning and achieves highly reliable grasp position selection by fully leveraging the interpretability of LLMs through prompt optimization and ontological reasoning. The sequential strategy—from task interpretation to part recognition, and finally to grasp selection—also demonstrates better stability compared to methods that integrate all features into a single embedding space.

\subsubsection{Evaluation of Grasp Success Rate}
Finally, we evaluate the reliability of part-aware grasp generation by comparing our method with GraspNet \cite{Fang2}, a grasp detection benchmark trained on large-scale datasets, and HGGD \cite{Siang}, a state-of-the-art grasp planning framework that outperforms GraspNet in dynamic and challenging scenarios. For both GraspNet and HGGD, we capture RGB-D images of the scene and use their pre-trained baseline models to generate grasp candidates with associated quality scores. Since these grasp detection methods are not capable of interpreting human instructions or recognizing functional object parts, this experiment focuses solely on grasp generation performance, assuming the target object parts are predefined. Given a specific object part, we apply our template-assisted approach to identify the best-matching point cluster $o^*_{part}$ within the observed point cloud. For all three methods, we select only grasp poses located within the region of $o^*_{part}$ for execution. In terms of grasp selection strategy, GraspNet and HGGD rank candidates by their predicted quality scores and select the one with the highest score, while our method selects the first candidate that successfully passes the stability-aware adjustment process. A grasp attempt is considered successful if the object is grasped at the designated functional part and lifted without being dropped. Failure cases include the absence of an executable grasp, an incorrect grasp location, or the object dropping during lifting. Each functional part of each object is subjected to 5 grasp attempts, with the object placed in different positions and orientations. In cases of part recognition failure, the trial is discarded and repeated with a new object pose. The final grasp success rate (GSR) is calculated as:
\begin{equation*}
    \mathrm{GSR= \frac{Number\;of\;successful\;grasps}{Number\;of\;total\;grasping\;attempts}}
\end{equation*}

\begin{table}[t]
\small
\renewcommand\arraystretch{1.5}
\setlength\tabcolsep{3pt}
\centering
\begin{tabular}{c | c c | c c | c c | c}
\toprule
Object Class & \multicolumn{2}{c|}{Mug} & \multicolumn{2}{c|}{Bottle} & \multicolumn{2}{c|}{Scissor} & \multirow{2}{*}{Average GSR} \\ 
Target part & Handle & \makecell{Body\\(Outside)} & Cap & Body & Handle & Blade \\
\hline
GraspNet & 3/15 & 10/15 & 3/15 & 12/15 & 0/15 & 2/15 & 30/90 (33.3\%) \\
HGGD & 3/15 & 12/15 & 5/15 & 14/15 & 0/15 & 0/15 & 34/90 (37.8\%) \\ 
Our method & 12/15 & 15/15 & 12/15 & 14/15 & 13/15 & 10/15 & \textbf{76/90 (84.4\%)} \\
\bottomrule
\end{tabular}
\caption{Evaluation of GSR via a Baseline Study Against Grasp Detection Benchmarks}\label{tab:3}
\end{table}

\begin{figure}[t]
    \centering
    \includegraphics[width=0.9\linewidth]{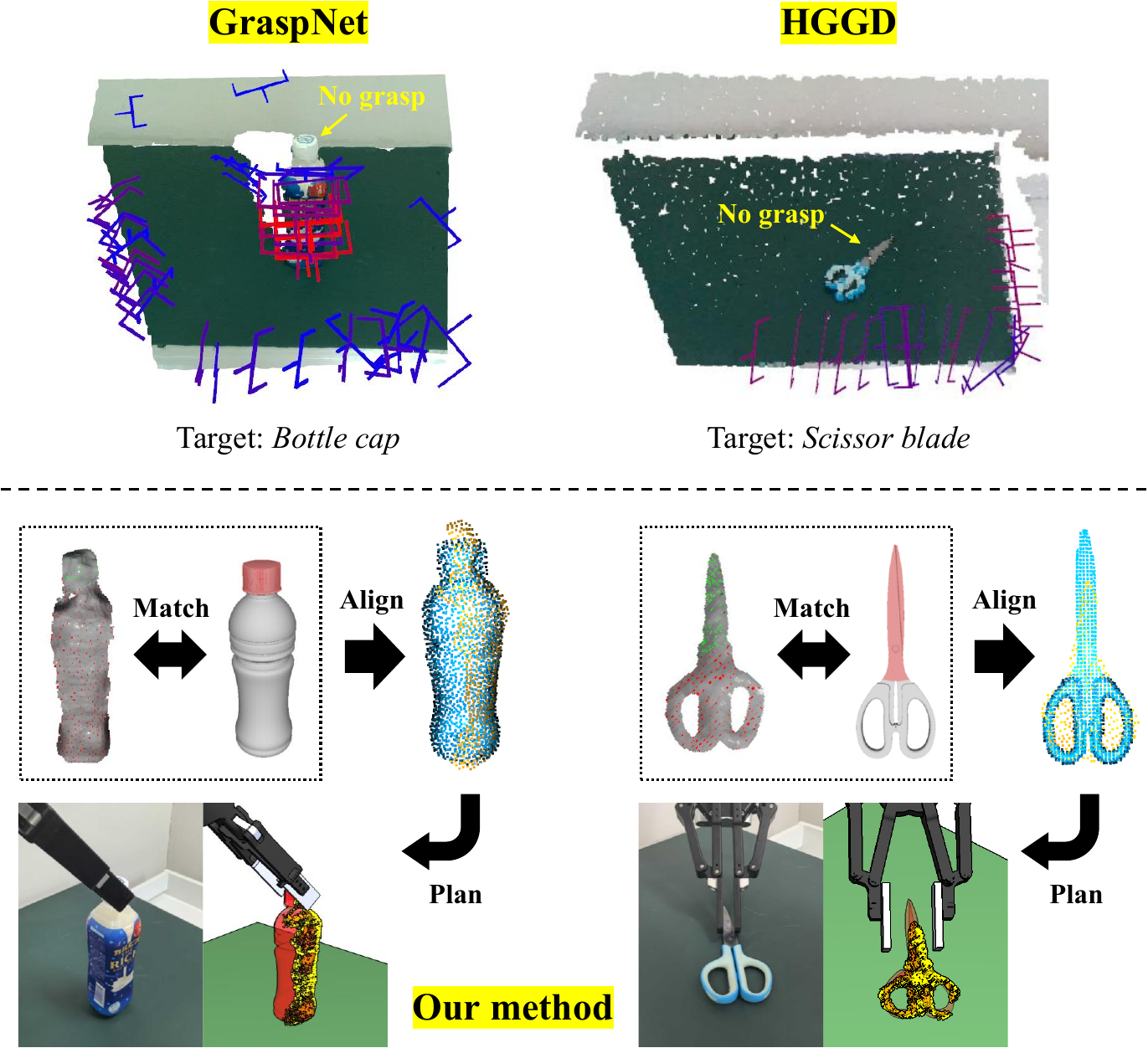}
    \caption{Comparison of TOG performance between benchmarks and our method.}
    \label{img:13}
\end{figure}

The experimental results are presented in Table \ref{tab:3}. Our method significantly outperforms both GraspNet and HGGD in grasping specific parts across various objects. We observe that GraspNet struggles to generate grasp candidates for small object parts, such as the \textit{bottle cap}, due to its tendency to favor larger surface areas that are more likely to ensure grasp robustness, as shown in Fig. \ref{img:13}. While HGGD demonstrates higher stability when grasping mugs and bottles, it fails to generate feasible grasps on either part of thin-shaped objects such as the \textit{scissor} in all trials, highlighting a key limitation of task-agnostic grasp detection approaches when applied to TOG problems. In contrast, our method consistently achieves high GSR across diverse object parts by leveraging similarity-based grasp planning combined with a localized positional adjustment process to optimize grasp performance.

\subsection{Ablation Study}
In the second TOG experiment, we further verify our local-to-global point cloud registration and stability-aware grasp adjustment through two baseline comparisons: \textit{direct registration} and \textit{no grasp adjustment}. In the \textit{direct registration} baseline, we employ a conventional point cloud registration pipeline using RANSAC followed by ICP to align the partial object point cloud with the complete template. In the \textit{no grasp adjustment} baseline, we execute the first generated grasp that is both IK-solvable and collision-free without considering the position of its grasp center. To clearly illustrate differences in grasp quality, we reuse the mugs from previous experiments, fill them with water, and designate their handles as the grasping targets. Each mug undergoes five grasp planning attempts under each baseline condition. Two types of failure are considered: 1) \textit{unsuccessful planning}, defined as either the absence of a generated grasp or a grasp targeting an incorrect object part, and 2) \textit{unstable grasping}, identified by water spilling from the mug during execution. For performance evaluation, we define the following two metrics:
\begin{align*}
    &\mathrm{PR\;(Planning\;Rate) = \frac{Number\;of\;successfully\;planned\;grasps}{Number\;of\;total\;grasp\;planning\;attempts}} \\[0.5em]
    &\mathrm{SR\;(Stabilization\;Rate) = \frac{Number\;of\;stable\;grasps}{Number\;of\;executed\;grasps}}
\end{align*}

\begin{table}[t]
\small
\renewcommand\arraystretch{1.5}
\setlength\tabcolsep{10pt}
\centering
\begin{tabular}{c | c c }
\toprule
Indicator & PR $\uparrow$ & SR $\uparrow$ \\ 
\hline
Direct registration & 33.3\% (5/15) & 60.0\% (3/5) \\
No grasp adjustment & 80.0\% (12/15) & 50.0\% (6/12) \\
Full method & \textbf{86.7\% (13/15)} & \textbf{76.9\% (10/13)} \\
\bottomrule
\end{tabular}
\caption{Ablation Study on Proposed Methods Using Water-Filled Mugs}\label{tab:4}
\end{table}

\begin{figure}[t]
    \centering
    \includegraphics[width=\linewidth]{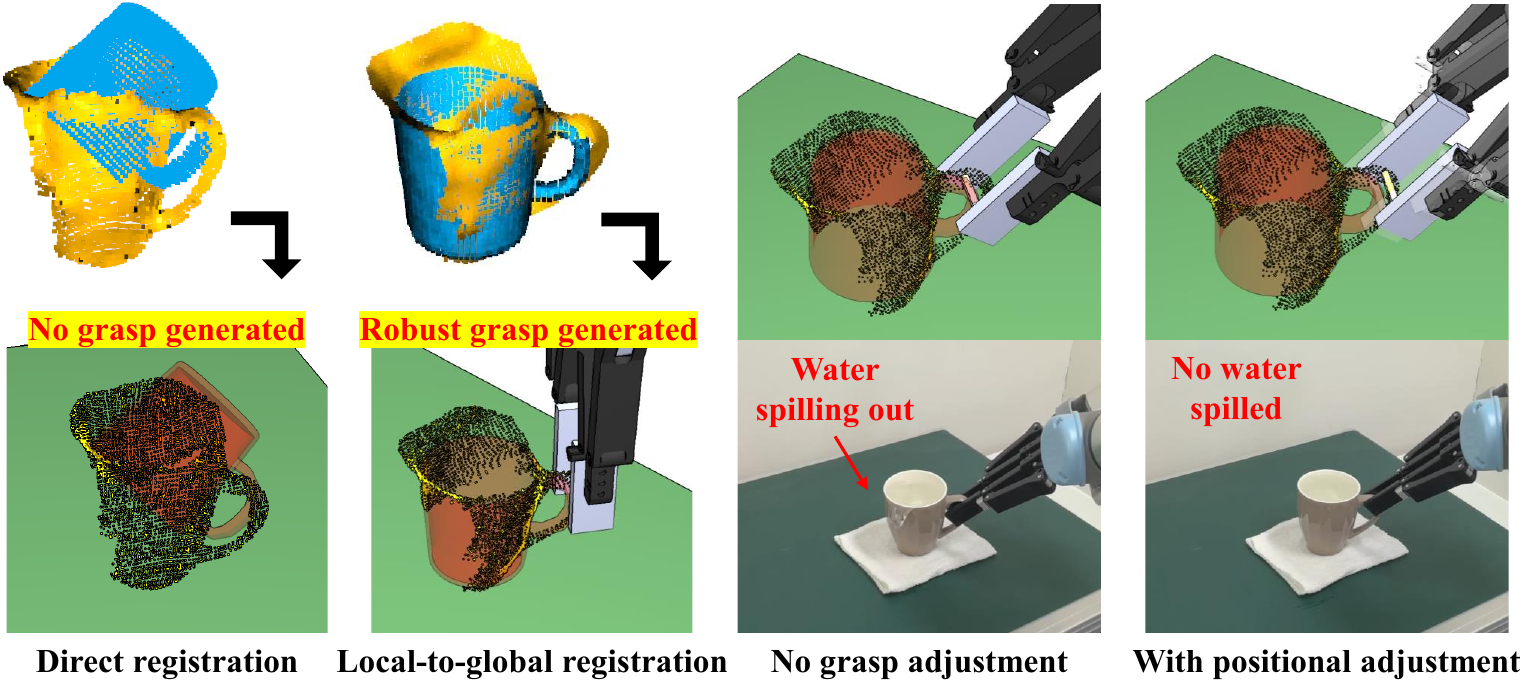}
    \caption{Performance discrepancy between the two baselines and our full method.}
    \label{img:14}
\end{figure}

In the case of \textit{unsuccessful planning}, the trial is terminated without executing any grasp and excluded from the calculation of SR. The experimental results are presented in Table \ref{tab:4}. Our full method significantly outperforms both baseline approaches across both evaluation metrics. The notable performance gap between the \textit{direct registration} baseline and our full method highlights the effectiveness of the proposed local-to-global point cloud registration in achieving precise matching between observed objects and model templates, which is an essential prerequisite for the reliable transfer of grasping knowledge and robust grasp generation. Additionally, the substantial improvement in SR observed between the \textit{no grasp adjustment} baseline and our full method underscores the importance of the stability-aware grasp adjustment in enhancing the quality of the executed grasps. A visual comparison showing the superiority of our full method is provided in Fig. \ref{img:14}. 

\subsection{Robustness to Occlusion, Noise, and Scale Variation}
\begin{figure}[t]
    \centering
    \includegraphics[width=\linewidth]{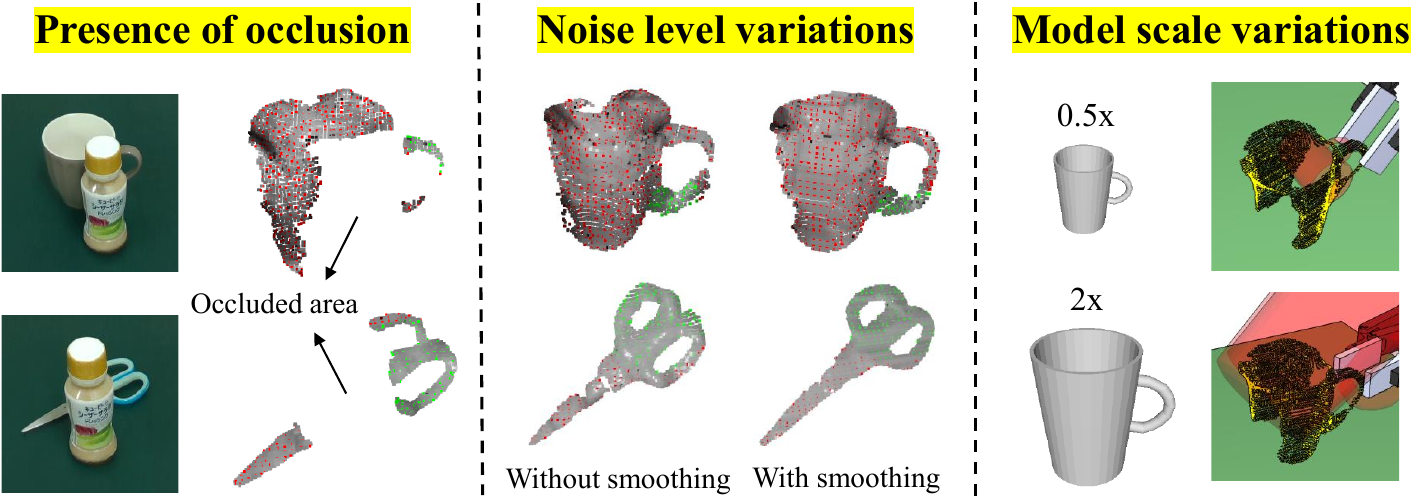}
    \caption{Robust part recognition and grasp planning under unstructured conditions.}
    \label{img:15}
\end{figure}

In the third TOG experiment, we evaluate the robustness of our proposed framework under various real-world uncertainties, focusing on three unstructured conditions: I) partial occlusion of the target object, II) sensor-induced noise in the point cloud, and III) scaling of model templates to varying sizes. The impacts of these conditions are shown in Fig. \ref{img:15}. Similar to Experiment \ref{exp}, we select the \textit{mug handle}, \textit{bottle cap}, and \textit{scissor handle} as representative functional parts for clear performance assessment.

For Condition I, we use a sauce bottle to block part of the target object within the camera view, resulting in a point cloud with substantial missing regions. Despite the sparsity and incompleteness of the visual input, our framework reliably identifies the functional parts, as shown in Fig. \ref{img:15} (left). This resilience is largely attributed to the use of multi-metric similarity evaluation, which enables robust recognition under partial observation. However, it is worth noting that recognition failures in extreme cases—such as when the target part is fully occluded—are beyond the scope of this study.

For Condition II, we employ the camera’s built-in filtering capabilities to simulate varying levels of sensor noise. Specifically, we apply the spatial filter provided by the RealSense SDK to smooth the original object point cloud. This filtering process significantly alters the point distribution, leading to noticeable variations in the object's geometry. Nevertheless, we observe that such variations have minimal impact on the accuracy of functional part recognition when employing the proposed sampling-based strategy, demonstrating the robustness of our method to sensor-induced noise.

For Condition III, we scale the model templates to both smaller and larger sizes to evaluate their matching and grasp planning performance on the same target object. Although changes in model size degrade the quality of point cloud matching (as shown in \ref{img:15}, right), our framework continues to generate feasible grasps on the functional parts. This robustness is achieved through a combination of local-to-global registration and stability-aware positional adjustment, which ensures that grasps transferred from the model templates are consistently positioned in appropriate and stable regions.

For further quantitative evaluation, we conduct 15 trials (3 different objects $\times$ 5 grasp attempts per object) for each selected object part under each unstructured condition, and compare the performance against that under the original condition. To provide a comprehensive assessment, we introduce a new evaluation metric, PGSR (Plan and Grasp Success Rate), defined as:
\begin{equation*}
    \mathrm{PGSR= \frac{Number\;of\;accurately\;planned\;and\;successfully\;executed\;grasps}{Number\;of\;total\;planning\;and\;grasping\;attempts}}
\end{equation*}
where only grasps planned on the accurate functional part and that stably lift the object are considered as successful. The results presented in Table \ref{tab:5} demonstrate the exceptional robustness of our proposed method under various real-world uncertainties. Each component of our framework—matching, registration, and adjustment—works in concert to form a reliable and effective solution for task-oriented object grasping.

\begin{table}[t]
\small
\renewcommand\arraystretch{1.5}
\setlength\tabcolsep{12pt}
\centering
\begin{tabular}{c | c c c | c}
\toprule
Object Class & Mug & Bottle & Scissor & \multirow{2}{*}{Average PGSR} \\ 
Functional part & Handle & Cap & Handle \\
\hline
Original condition & \textbf{11/15} & \textbf{12/15} & \textbf{13/15} & \textbf{36/45 (80.0\%)} \\
With occlusion & 9/15 & 10/15 & 13/15 & 32/45 (71.1\%) \\
With noise & 10/15 & 12/15 & 13/15 & 35/45 (77.8\%) \\
With scaling & 9/15 & 12/15 & 12/15 & 33/45 (73.3\%) \\
\bottomrule
\end{tabular}
\caption{Quantitative Evaluation of Performance Robustness Under Varying Conditions}\label{tab:5}
\end{table}

However, it is important to acknowledge that the robustness of our method is limited under certain levels of interference. More challenging sensing conditions—such as intense illumination, object transparency, or highly cluttered scenes—that cause significant changes in visual features are beyond the scope of this study. Combining our approach with point cloud refinement or shape reconstruction techniques, which can recover object geometry from noisy or distorted features, may offer a potential solution for these rigorous conditions.

\subsection{Generalization to Novel-Category Objects}\label{gen}
Finally, in addition to delivering high performance on objects included in the ontology, our method demonstrates strong generalizability to novel-category objects, extending beyond the existing knowledge base. This is accomplished by leveraging the scalable inferential capabilities of LLMs, which can be activated by appending the following instruction to the prompt:

\textit{
If the target object is not listed in the ontology, find its closest object in the ontology and use its part information.
}

\begin{figure}[t]
    \centering
    \includegraphics[width=\linewidth]{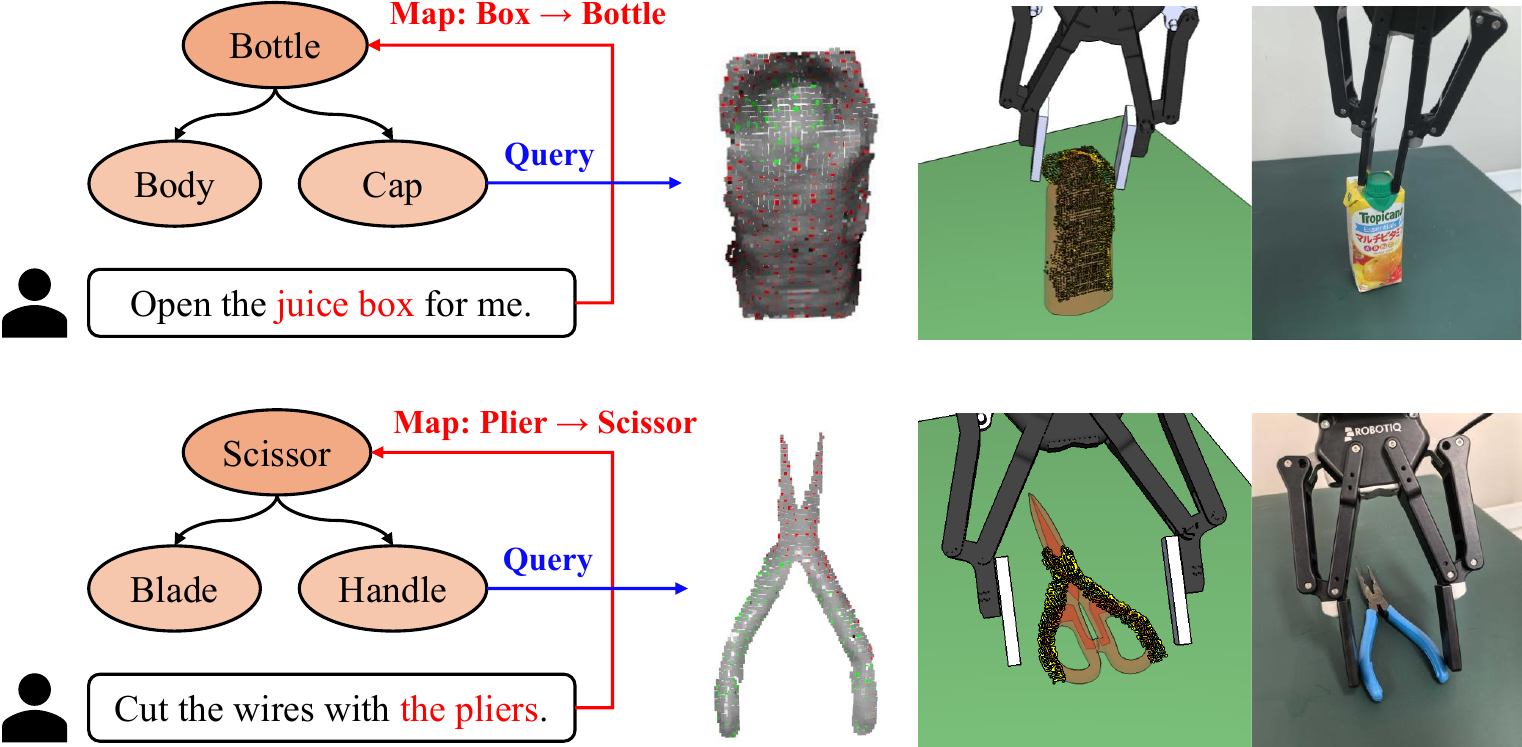}
    \caption{Generalizing TOG to novel categories via ontological knowledge extension.}
    \label{img:16}
\end{figure}

The answer template is also modified accordingly, as described in \ref{B}. Through this prompt refinement, a novel target object can be mapped to a similar known instance within the ontology, allowing the reuse of both ontological knowledge and corresponding model templates for subsequent recognition, matching, and planning. Fig. \ref{img:16} illustrates two representative examples involving a \textit{juice box} and a pair of \textit{pliers}. Despite lacking prior knowledge of these objects, the system successfully maps them to the \textit{bottle} and \textit{scissor} categories in the ontology, respectively. Their subclasses and associated templates are then employed to guide functional part recognition and similarity-based grasp planning. From this perspective, our approach exhibits a tree-like growth pattern: each newly introduced instance contributes to the exponential expansion of existing knowledge. Compared to conventional model-based methods that encode diverse object features into a single network, our framework offers superior reliability and robustness in handling novel objects through structural and scalable ontological reasoning.

\begin{figure}[t]
    \centering
    \includegraphics[width=\linewidth]{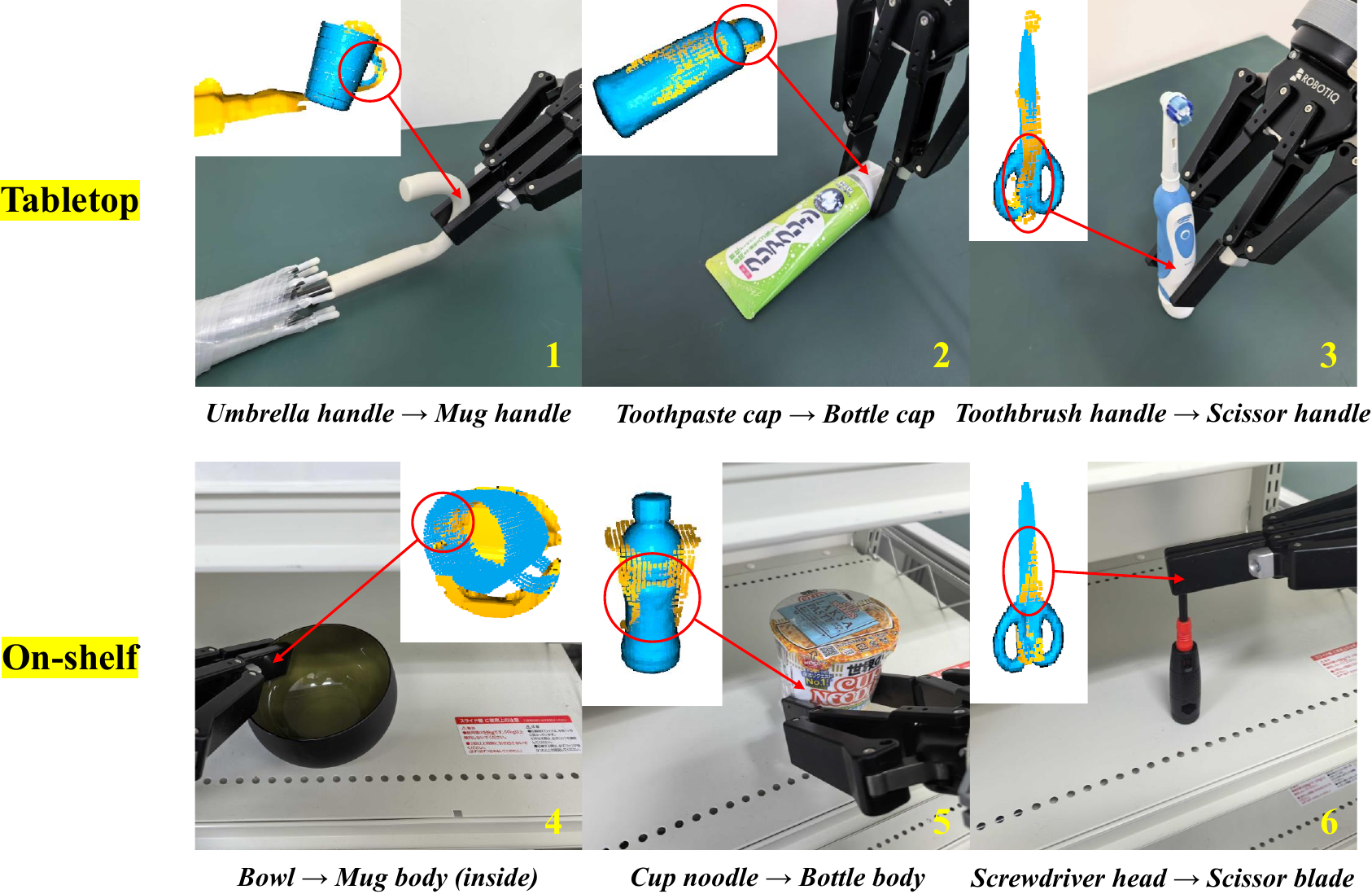}
    \caption{Validation on diverse novel objects across tabletop and on-shelf scenarios.}
    \label{img:17}
\end{figure}

For further quantitative evaluation, we perform TOG experiments on a diverse set of novel objects across two real-world settings: \textit{tabletop} and \textit{on-shelf}. As illustrated in Fig. \ref{img:17}, our method effectively maps unknown target functional parts to predefined object parts within the ontology, enabling the generation of task-oriented grasps for a wide variety of unseen objects that differ significantly from existing templates. Notably, the matching results shown in the top corners highlight the effectiveness of the local-to-global registration strategy in achieving accurate alignment between functional parts, even when the overall object geometry varies considerably.

Table \ref{tab:6} presents the overall experimental results. The object numbers correspond to those shown in Fig. \ref{img:17}. Each object is tested in 5 randomized positions and orientations, with the following task instructions used as inputs:

1) \textit{Bring me the umbrella.} (Target: umbrella handle)

2) \textit{Get the toothpaste ready for use.} (Target: toothpaste cap)

3) \textit{Pick up the toothbrush.} (Target: toothbrush handle)

4) \textit{Take the bowl off the shelf.} (Target: bowl)

5) \textit{Fetch the cup noodles for me to eat.} (Target: cup noodle)

6) \textit{Hand me the screwdriver so I can use it.} (Target: screwdriver head)

The results of PRA and GSR clearly demonstrate the effectiveness of our proposed method in handling a wide variety of novel objects across different operation locations. By focusing on the similarity between local functional parts rather than the global object geometry, our approach achieves extensive generalization capacity despite a limited number of existing templates.

\begin{table}[t]
\small
\renewcommand\arraystretch{1.5}
\setlength\tabcolsep{10pt}
\centering
\begin{tabular}{c | c c c c c c | c}
\toprule
Object No. & 1 & 2 & 3 & 4 & 5 & 6 & Average \\
\hline
PRA & 4/5 & 5/5 & 4/5 & 5/5 & 5/5 & 4/5 & 27/30 (90.0\%) \\
GSR & 4/4 & 4/5 & 4/4 & 3/5 & 5/5 & 3/4 & 23/27 (85.2\%) \\
\bottomrule
\end{tabular}
\caption{Quantitative Evaluation of Generalization Performance on Diverse Novel Objects}\label{tab:6}
\end{table}

\section{Discussion}
\subsection{Strengths and Limitations}
The experiments above demonstrate the superiority of our proposed method in leveraging a small number of templates to efficiently handle a wide range of novel objects. This strong generalizability primarily stems from emphasizing \textit{local part similarity} rather than \textit{global object similarity}. While previous studies have explored part registration for grasp transfer from demonstrated to unseen objects \cite{Tekden,Wu2,Simeonov}, their approaches typically require transforming part features into implicit representations for neural network training, followed by retrieving grasp poses from an embedding space. Such training pipelines reduce the interpretability of grasp planning outcomes—particularly in failure cases—thereby hindering model refinement and generalization. In contrast, our method directly aligns object parts using explicit geometric features through a local-to-global registration strategy. This design ensures traceability of grasp transfer errors and facilitates the optimal selection of model templates with minimal transfer error, leading to greater performance stability across diverse object categories.

\begin{figure}[t]
    \centering
    \includegraphics[width=\linewidth]{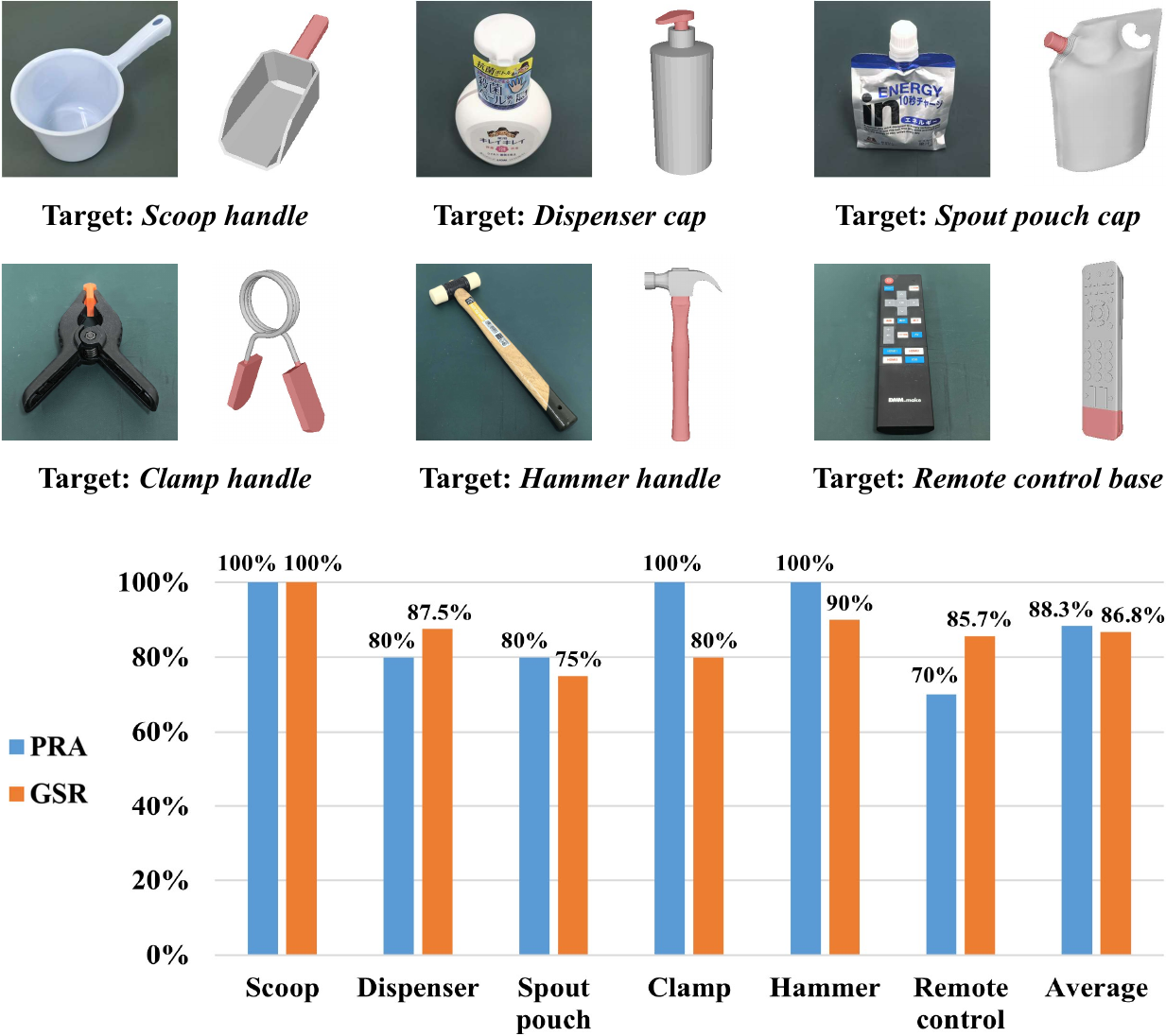}
    \caption{Extended evaluation across a broader set of object categories and tasks.}
    \label{img:18}
\end{figure}

To further evaluate the strengths and limitations of our proposed method, we expand the initial experimental set from three object categories to a broader set, following existing TOG benchmarks \cite{Ju,Rashid,Tang2}. The upper part of Fig.~\ref{img:18} illustrates the selected objects, their corresponding model templates, and the targeted functional parts. As in the previous experiments, we prepared three templates per object (only one is shown in the figure), all within the same category but varying in geometry. Each object was tested 10 times for both part recognition and grasp planning, using the following task instructions as inputs:

1) \textit{Use the scoop to get some water for me.} (Target: Scoop handle)

2) \textit{Press the dispenser to release soap.} (Target: Dispenser cap)

3) \textit{Unscrew the spout pouch so I can drink.} (Target: Spout pouch cap)

4) \textit{Open the clamp to secure the books.} (Target: Clamp handle)

5) \textit{Pass me the hammer.} (Target: Hammer handle)

6) \textit{Hold the remote control steady.} (Target: Remote control base)

The results in the lower part of Fig. \ref{img:18} demonstrate the consistent superiority of our method across diverse object categories. Exploiting the local similarity of functional parts provides sufficient accuracy and stability when matching unseen target objects to semantically similar templates, even when their global geometries differ. However, failures occur when the target part is small with sparse features (e.g., the spout pouch cap) or has a geometrically similar counterpart on the symmetrical side (e.g., the remote control base). In such cases, the part’s geometric features are less distinctive relative to the entire object, and similar regions on the surface can significantly interfere with part recognition. This represents a potential limitation of the current approach when dealing with specific object types.

\subsection{Guidelines for Database Construction}
In addition to the registration strategy, constructing an appropriate template database is a critical step in achieving optimal performance. While our ultimate goal is to generalize the method to handle open-ended objects and tasks, several key considerations must be addressed: I) What types of templates should be included in the database? II) How many instances should be collected for each object category? III) What intra-category variation should be considered during data collection? Based on our experimental results, we summarize the following principles for selecting model templates:

I) In the context of robotic grasping, typical functional parts can be categorized as \textit{Graspable Parts} (e.g., handle, body), \textit{Active Parts} (e.g., blade, head), \textit{Support Parts} (e.g., bottom, base), and \textit{Container Parts} (e.g., cap, lid). For each part type, one representative object category can be included in the database and generalized to other categories with similar functional parts. For example, incorporating a \textit{mug} template with the functional part \textit{handle} enables the method to generalize to \textit{pots}, \textit{kettles}, and \textit{umbrellas} that share similar handles. However, a single category may not fully represent all variations—for instance, the \textit{scissor handle} and \textit{hammer handle} differ~significantly from a \textit{mug handle} in geometry. Therefore, template selection should consider high-level category associations (e.g., \textit{handle} appearing in both \textit{containers} and \textit{tools}) to ensure comprehensive coverage of object types.

II) Regarding the number of templates per object category, our TOG experiments (Fig. \ref{img:10}) show that three templates with different shapes and sizes are generally sufficient for good performance. Nevertheless, the optimal number should be adaptively determined based on category-specific requirements, which calls for a systematic derivation of the relationship between template quantity and object characteristics. We leave this derivation to future work for a more precise determination of category-aware template quantities.

III) Within each object category, the selected templates should exhibit diversity in shape, size, and texture to capture a broad range of real-world instances. To minimize database redundancy, templates with overly similar properties should be avoided. While our method demonstrates robustness to object variations, real-world tests confirm that a more diverse template set increases the likelihood of finding relevant references, typically resulting in higher part-matching accuracy and improved grasp stability.

Based on these principles, a compact yet highly effective template database capable of supporting a wide range of objects and tasks can be constructed.

\subsection{Accuracy-Runtime Trade-off}
We further investigate the precise impact of a growing template database on TOG performance and computational cost, aiming to provide a valid reference for real-world deployment. To this end, we reuse the objects shown in Fig. \ref{img:18} and perform the same trials while varying the number of templates used for each object. The results of PRA, GSR, and runtime for the matching process are recorded in Table \ref{tab:7}. Consistent with the results in Fig. \ref{img:10}, both recognition and grasp accuracy improve as the number of templates increases from 1 to 3, but remain virtually unchanged thereafter. Although the best performance is observed with 10 $\sim$ 50 templates, the significant increase in runtime renders the marginal performance gains negligible. In conclusion, the number of templates per object category should be set around 5 in the initial stage, with further adjustments made based on intra-category variations.

\begin{table}[t]
\small
\renewcommand\arraystretch{1.5}
\setlength\tabcolsep{10pt}
\centering
\begin{tabular}{c | c c c c c}
\toprule
Number of Templates & 1 & 3 & 5 & 10 & 50 \\
\hline
PRA $\uparrow$ & 65.0\% & 88.3\% & 90.0\% & 90.0\% & \textbf{91.7\%} \\
GSR $\uparrow$ & 61.5\% & 86.8\% & 87.0\% & \textbf{88.9\%} & 87.3\% \\
Runtime $\downarrow$ & \textbf{0.35s} & 0.63s & 0.89s & 1.49s & 6.72s \\
\bottomrule
\end{tabular}
\caption{Quantitative Trade-off Analysis: Recognition, Grasping, and Computation}\label{tab:7}
\end{table}

\section{Conclusion}
In this paper, we present a novel strategy for task-oriented grasping of unseen objects. Our approach introduces an object-part-task ontology consisting of online and offline components, associated through guidance from LLMs. By leveraging optimized user prompts, the LLM accurately interprets intuitive human instructions and infers corresponding functional object parts based on ontological knowledge. Additionally, we propose a novel part recognition method that utilizes pre-existing model templates to identify target object parts as point clusters. Based on this representation, we employ a local-to-global point cloud registration framework, followed by a stability-aware grasp adjustment process, to transfer grasping knowledge from the best-matching model template to the unseen target object, enabling robust task-oriented grasp generation. The stable and efficient integration of task reasoning (language), object-part recognition (vision), and grasp detection (action) constitutes the main novelty of our work. To validate this, we first conduct real-world experiments on similar-category objects, demonstrating its superiority over state-of-the-art approaches. Furthermore, we highlight its scalability to a wide range of novel-category objects using a target-agnostic, pre-generated database composed of only a small set of model templates.

Despite its strengths, the proposed method has several limitations: 1) it cannot handle completely novel objects that lack semantically or geometrically similar references in the ontology; 2) its matching accuracy degrades when the viewpoint provides insufficient features of the target part or when the part is not geometrically distinctive; and 3) it does not currently account for constraints imposed by subsequent manipulation tasks beyond grasp execution. Addressing these limitations will be the focus of our future work.

\appendix
\section{LLM Prompt Optimization}
\label{A}
To optimize LLM prompts for accurate instruction interpretation, we adopt the framework proposed in \cite{Yang}, which consists of three components: 1) \textit{Executor}, the LLM that processes the given prompt and generates a corresponding response; 2) \textit{Evaluator}, a human supervisor who assesses the LLM's output and determines the next step—either terminating the optimization if the result is satisfactory, or providing feedback for improvement if not; and 3) \textit{Improver}, where the LLM is reused to refine the prompt based on human feedback. The improved prompt is then fed back into the \textit{Evaluator}, initiating a new iteration of the process (see Fig. \ref{img:19}). The key advantage of this framework is that humans only need to act as advisors, offering suggestions rather than revising the prompts themselves. The LLM (essentially a more efficient “engineer”), through repeated iterations and human supervision, ultimately generates an optimal prompt. Below, we illustrate an example of prompt optimization for instructing a TOG task:

\begin{figure}[t]
    \centering
    \includegraphics[width=0.7\linewidth]{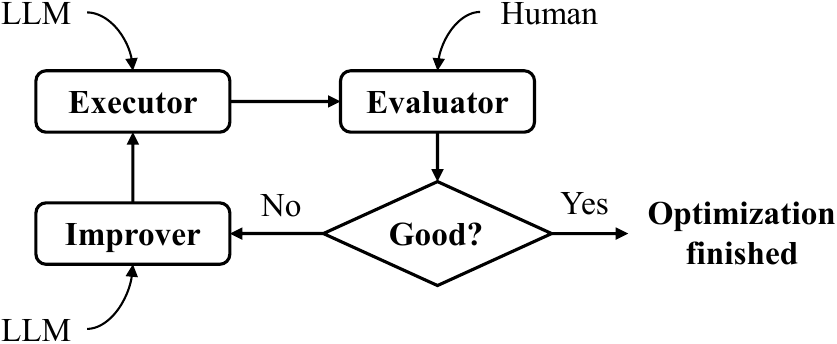}
    \caption{Framework for reusing LLMs in iterative prompt optimization.}
    \label{img:19}
\end{figure}

\vspace{10pt}

\newcommand{\user}[1]{\textbf{User:} \textit{#1}\\}
\newcommand{\agent}[1]{\textbf{Agent:} \textit{#1}\\}
\textbf{Create Agent 1}
\begin{quote}
\user{
Given the following ontology: \\
Mug → Handle, Mug → Body → Inside, Mug → Body → Outside \\
Scissor → Handle, Scissor → Blade \\
Considering these two commands given to a robot: \\
1. "Hand me the mug with hot water in it." \\
2. "Pick up the scissor and pass it to me. \\
Question: Which part(s) of each object should the robot grasp? Please only answer the part names.}

\agent{1. Mug → Handle 2. Scissor → Handle}
(Ground truth) \underline{1. Mug → Body → Outside
2. Scissor → Blade}

\user{Both the answers are wrong because in a human-robot interaction task, we want the robot to handle dangerous parts instead of the human. Meanwhile, the robot and human should take different parts to ensure sufficient operating space. How can we improve the prompt to obtain correct answers?}

\agent{You can improve the question by explicitly stating the criteria for grasping ... (Unimportant statements) \\
Revised Prompt: ``Given the following ontology: \\
Mug → Handle, Mug → Body → Inside, Mug → Body → Outside \\
Scissor → Handle, Scissor → Blade \\
Consider these two commands given to a robot: \\
1. Hand me the mug with hot water in it \\
2. Pick up the scissor and pass it to me \\
Which part(s) of the objects should the robot grasp, ensuring that (a) the robot handles dangerous parts instead of the human, and (b) the robot and human grasp different parts for sufficient operating space?" \\
This revision ensures that the intended grasping strategy is clear, leading to correct answers ... (Unimportant statements)}
\end{quote}

\textbf{Create Agent 2}
\begin{quote}
\user{(The revised prompt from Agent 1)}

\agent{1. Mug → Handle 2. Scissor → Blade}
(Ground truth) \underline{1. Mug → Body → Outside
2. Scissor → Blade}

\user{The first answer is still wrong because in a human-robot interaction task, we want the robot to handle dangerous parts instead of the human. Meanwhile, the robot and human should take different parts to ensure sufficient operating space. How can we improve the prompt to obtain correct answers?}

\agent{To ensure correct answers, we should explicitly state the constraints more clearly ... (Unimportant statements)\\
Revised Prompt: ``Given the following ontology: \\
Mug → Handle, Mug → Body → Inside, Mug → Body → Outside \\
Scissor → Handle, Scissor → Blade \\
A robot is given the following commands: \\
1. Hand me the mug with hot water in it \\
2. Pick up the scissor and pass it to me \\
Constraints: \\
1. The robot should grasp parts that are either difficult to manipulate or potentially dangerous, while the human should grasp the safer, easier-to-handle parts. \\
2. The robot and human should each grasp a different part of the object to ensure enough operating space. \\
Question: Which part(s) of each object should the robot grasp? Answer with only the part names." \\
This version explicitly defines the constraints, making it clear that ... (Unimportant statements)}
\end{quote}

\textbf{Create Agent 3}
\begin{quote}
\user{(The revised prompt from Agent 2)}

\agent{1. Mug → Body → Outside 2. Scissor → Blade}
(Ground truth) \underline{1. Mug → Body → Outside
2. Scissor → Blade}
\end{quote}

\textbf{Prompt optimization completed}

\section{Modified Answer Template}
\label{B}
To fully leverage the scalability of LLMs for novel-category objects, we design the following step-by-step reasoning process as the answer template:

\newcommand{\blank}[1]{\textit{#1}\\}
\begin{quote}
\blank{The given command is ``Empty the bowl into the sink." \\
\textbf{Step 1: Identify the Type of Task} \\
The robot is instructed to pick up the bowl and pour its contents into the sink. \\
There is no human involved in handling the bowl. \\
Therefore, this is a manipulation task. \\
\textbf{Step 2: Find the Closest Object in the Ontology} \\
The bowl is not in the ontology. Let’s look for the most structurally similar object: \\
The bowl, like the mug, is a container with a concave inside and round outer body. \\
The key difference: a mug typically has a handle, whereas a bowl usually does not. \\
Despite this, the bowl's structure (body with inside and outside) closely aligns with the mug’s body parts. \\
\textbf{So, we map: Bowl $\approx$ Mug (without handle)} \\
Relevant parts: \\
Body → inside \\
Body → outside \\
\textbf{Step 3: Apply Task Constraints} \\
The robot must grasp the bowl in a way that allows it to tilt and pour the contents into the sink. \\
This requires a secure and controlled grip—likely from the outside of the body to allow for rotation. \\
\textbf{Analyzing the Object Parts} \\
Body (inside): Contains the contents—not suitable for gripping. \\
Body (outside): Provides surface area for a stable grip and control when tilting. \\
\textbf{Best Choice for the Robot} \\
The robot should grasp the body (outside) of the bowl to securely hold, tilt, and empty its contents into the sink. \\
\textbf{Conclusion}: The Robot Should Grasp the Body (Outside) \\
This enables the robot to effectively manipulate the bowl for pouring while maintaining control and avoiding contact with the contents.}
\end{quote}


\clearpage
\section*{Declaration of generative AI and AI-assisted technologies in the writing process}
During the preparation of this work the author(s) used ChatGPT in order to improve the readability and language of the manuscript. After using this tool/service, the author(s) reviewed and edited the content as needed and take(s) full responsibility for the content of the published article.

\section*{Acknowledgment}
This research is subsidized by New Energy and Industrial Technology Development Organization (NEDO) under a project JPNP20016. This paper is one of the achievements of joint research with and is jointly owned copyrighted material of ROBOT Industrial Basic Technology Collaborative Innovation Partnership (ROBOCIP).

\bibliographystyle{elsarticle-num}
\bibliography{reference}






\end{document}